\documentclass{article}
\usepackage{ijcai25}

\usepackage{times}
\usepackage{soul}
\usepackage{url}
\usepackage[hidelinks]{hyperref}
\usepackage[utf8]{inputenc}
\usepackage[small]{caption}
\usepackage{graphicx}
\usepackage{amsmath}
\usepackage{amsthm}
\usepackage{booktabs}
\usepackage{algorithm}
\usepackage{algorithmic}
\usepackage[switch]{lineno}
\usepackage{multirow}
\usepackage{arydshln}
\usepackage{subcaption}
\usepackage{amssymb}
\usepackage{pifont}
\usepackage[table]{xcolor}

\title{Maximum Redundancy Pruning: A Principle-Driven Layerwise Sparsity Allocation for LLMs}

\author{
Chang Gao$^1$
\and
Kang Zhao$^2$\and
Runqi Wang$^1$\and
Jianfei Chen$^2$\And
Liping Jing$^1$\\
\affiliations
$^1$Beijing Jiaotong University\\
$^2$Tsinghua University\\
\emails
22110098@bjtu.edu.cn,
zhaokang29@huawei.com,
rqwang@bjtu.edu.cn,
jianfeic@tsinghua.edu.cn,
lpjing@bjtu.edu.cn
}

\begin{document}

\title{Maximum Redundancy Pruning: A Principle-Driven Layerwise Sparsity Allocation for LLMs}





\maketitle
\begin{abstract}
  Large language models (LLMs) have demonstrated impressive capabilities, but their enormous size poses significant challenges for deployment in real-world applications. To address this issue, researchers have sought to apply network pruning techniques to LLMs. A critical challenge in pruning is the allocation of sparsity for each layer. Recent sparsity allocation methods are often based on heuristics or search that can easily lead to suboptimal performance. In this paper, we conducted an extensive investigation into various LLMs and revealed three significant discoveries: (1) the Layerwise Pruning Sensitivity (LPS) of LLMs is highly non-uniform, (2) the choice of pruning metric affects LPS, and (3) the performance of a sparse model is related to the uniformity of its layerwise redundancy level. Based on these discoveries, we propose that the layerwise sparsity of LLMs should adhere to three principles: \emph{non-uniformity}, \emph{pruning metric dependency}, and \emph{uniform layerwise redundancy level} in the pruned model. To this end, we proposed Maximum Redundancy Pruning (MRP), an iterative pruning algorithm that prunes in the most redundant layers (\emph{i.e.}, those with the highest non-outlier ratio) at each iteration. The achieved layerwise sparsity aligns with the outlined principles. We conducted extensive experiments on publicly available LLMs, including LLaMA2 and OPT, on various benchmarks. The experimental results validate the effectiveness of MRP, demonstrating its superiority over previous methods. We make our code available at \url{https://github.com/Gaochang-bjtu/MRP}.
\end{abstract}
\section{Introduction}
\label{sec1}

Large Language Models (LLMs) have exhibited remarkable capabilities across various applications \cite{bubeck2023sparks,chowdhery2023palm}, which in turn has motivated researchers to explore strategies for their deployment \cite{luccioni2023estimating,patterson2021carbon}. However, their massive size and high computational demands raise concerns about financial costs and environmental impact. Consequently, efficiently compressing LLMs has become a key research focus.\par

Network pruning \cite{hassibi1993optimal}, a well-established model compression technique, holds promise as a solution for reducing the size of LLMs. Several approaches, such as SparseGPT \cite{frantar2023sparsegpt} and Wanda \cite{sun2024pruning}, have been specifically developed for LLM pruning. SparseGPT
prunes unimportant weights and reconstructs layerwise outputs. Wanda prunes weights based on the product of weights and activation magnitudes. These methods assign a uniform sparsity to each layer, which is often suboptimal given the varying importance of layers. Recent methods employ search strategies (e.g., evolutionary algorithms \cite{li2024discovering} and linear programming \cite{li2024adaptive}) or heuristic functions \cite{yin2024outlier} to allocate layerwise sparsity. However, for high-dimensional search spaces, these approaches often yield suboptimal solutions. This dilemma highlights the challenge of finding an optimal layerwise sparsity in a high-dimensional space without principled constraints. Consequently, a pressing question arises: 
\begin{center}
\emph{What principles should layerwise sparsity follow for LLMs? }
\end{center}
\par
\begin{figure}[t]

\centering

\begin{minipage}[b]{0.24\textwidth}
  \centering
  \includegraphics[width=0.95\textwidth]{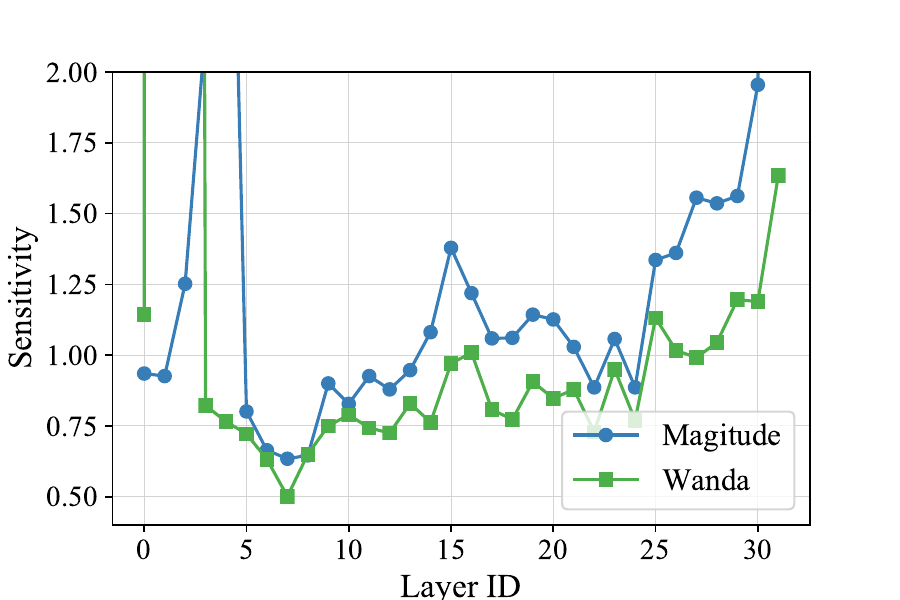}
  \label{in:1}
  
\end{minipage}
\hfill
\begin{minipage}[b]{0.22\textwidth}
  \centering
  \includegraphics[width=0.95\textwidth]{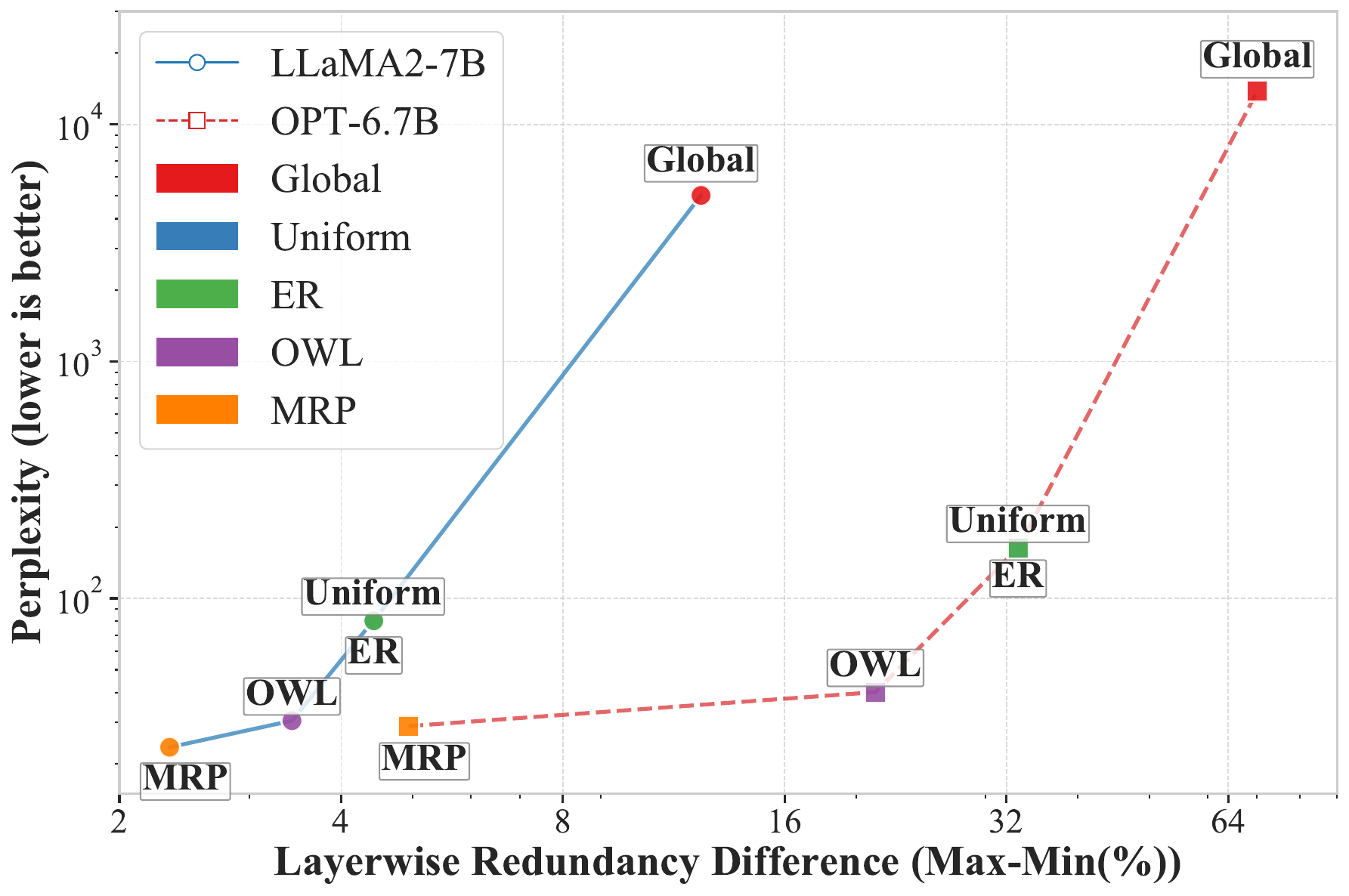}
  \label{in:2}
  
\end{minipage}
\vspace{-5pt}
\caption{The left: The LPS (increase in WikiText2 perplexity) of the OPT-6.7B under different pruning metrics. The right: WikiText2 perplexity and layerwise redundancy differences of different sparsity allocations (pruning metric: Wanda).}
\label{in}
\vspace{-17pt}
\end{figure}
To answer this question, we analyze the sensitivity of each layer in LLMs to pruning. Based on empirical results under various settings—including different architectures, pruning metrics, and model sizes—we draw several key conclusions. First, the sensitivity of different layers to pruning varies significantly: some layers maintain performance even under high sparsity, while others experience severe degradation. This observation suggests that LLM pruning should adopt \emph{non-uniform} layerwise sparsity. Second, we find that layer sensitivity is influenced by the pruning metric (as shown in Fig. \ref{in}: left), indicating that LLM pruning should use the \emph{metric-dependent} layerwise sparsity. Third, the performance of sparse models is correlated with the uniformity of layerwise redundancy (as shown in Fig. \ref{in}: right), suggesting that sparsity allocation should aim to \emph{make layerwise redundancy more uniform}. \par
Directly motivated by these empirical findings, we propose an iterative Maximum Redundancy Pruning (MRP) method that satisfies all three principles: it (1) applies non-uniform sparsity across layers, (2) considers the impact of pruning metrics on sparsity allocation, and (3) iteratively prunes the most redundant layers to balance the overall redundancy distribution. In each iteration, this method quantifies layerwise redundancy using non-outlier ratios and prunes in the most redundant layer until the target sparsity is achieved. \par

We conducted a comprehensive empirical evaluation to assess the generalizability of MRP in LLM architectures, including LLaMA2 and OPT. Our extensive experiments demonstrate that MRP consistently outperforms state-of-the-art LLM pruning methods in both language modeling and zero-shot classification tasks. MRP also excels when applied to vision and multimodal models (DeiT, ConvNeXt, LLaVA), demonstrating its versatility across architectures and modalities. In particular, the LLaMA2-13B model pruned to 7B parameters using "MRP+LoRA" slightly outperforms the LLaMA2-7B model trained from scratch, marking a significant breakthrough in efficient model compression.
 \begin{table*}[t]
        
	\centering
        \renewcommand{\arraystretch}{0.95}
        \caption{The summary of LPS across sparsity levels. Only the maximum and minimum values are shown here.  WikiText2 perplexity for LLMs; MM-Vet results for LLaVA at 70\% sparsity (\textbf{L}anguage and \textbf{V}ision heads separately).}
	\begin{tabular}{c  c  c c c c cc cc c}
		\toprule
		\multirow{2}{*}{Model} & \multirow{2}{*}{Param} & \multirow{2}{*}{Granularity} & \multirow{2}{*}{Metric}  & \multicolumn{2}{c}{60\% / 4:8} & \multicolumn{2}{c}{70\% / 5:8} & \multicolumn{2}{c}{80\% / 6:8} &  \multirow{2}{*}{Non-uniform}\\
        &&&&Min & Max & Min & Max & Min & Max   &\\
        \midrule
        \multirow{10}{*}{LLaMA2} & \multirow{5}{*}{7B} & \multirow{2}{*}{Unstructured} & Magnitude  & 0.05 & 3.80 & 0.09 & 5.41 & 0.15 & 16.00 & \textcolor{green}{\ding{51}}\\
        & && Wanda  & 0.01 & 0.20 & 0.04 & 0.55 & 0.09 & 1.34 & \textcolor{green}{\ding{51}}\\
        \cline{3-11}
        & & \multirow{2}{*}{Semi-structured} & Magnitude  & 0.03 & 1.54 & 0.07 & 2.68 & 0.15 & 19.20 & \textcolor{green}{\ding{51}}\\
        & & & Wanda  & 0.01 & 0.15 & 0.03 & 0.49 & 0.10 & 1.34 & \textcolor{green}{\ding{51}}\\
        \cline{3-11}
        & & Structured & Wanda  & 0.24 & 6.8e2 & 0.23 & 3e3 & 0.22 & 2.1e4 & \textcolor{green}{\ding{51}}\\
        \cline{3-11}
         & \multirow{5}{*}{13B} & \multirow{2}{*}{Unstructured} & Magnitude  & 0.03 & 0.50 & 0.06 & 0.68 & 0.10 &0.90 & \textcolor{green}{\ding{51}}\\
        & && Wanda  & 0.01 & 0.14 & 0.03 & 0.31 & 0.06 & 0.66 & \textcolor{green}{\ding{51}}\\
        \cline{3-11}
        & & \multirow{2}{*}{Semi-structured} & Magnitude  & 0.03 & 0.32 & 0.05 & 0.59 & 0.09 & 1.77 & \textcolor{green}{\ding{51}}\\
        & & & Wanda  & 0.01 & 0.11 & 0.02 & 0.29 & 0.07 & 0.67 & \textcolor{green}{\ding{51}}\\
        \cline{3-11}
        & & Structured & Wanda  & 0.13 & 1.5e3 & 0.13 & 3e4 & 0.13 & 3.3e4 & \textcolor{green}{\ding{51}}\\
        \midrule
        \multirow{5}{*}{OPT} & \multirow{5}{*}{6.7B} & \multirow{2}{*}{Unstructured} & Magnitude  & 0.64 & 4.94 & 0.63 & 6.66 & 0.60 & 1.6e3 & \textcolor{green}{\ding{51}}\\
        & && Wanda  & 0.51 & 1.32 & 0.50 & 2.2e2 & 0.52 & 3.2e4 & \textcolor{green}{\ding{51}}\\
        \cline{3-11}
        & & \multirow{2}{*}{Semi-structured} & Magnitude  & 0.55 & 2.29 & 0.56 & 4.2e2 & 0.58 & 2.6e4 & \textcolor{green}{\ding{51}}\\
        & & & Wanda  & 0.61 & 1.22 & 0.56 & 6.92 & 0.55 & 3.1e4 & \textcolor{green}{\ding{51}}\\
        \cline{3-11}
        & & Structured & Wanda  & 0.51 & 20.58 & 0.49 & 1.7e2 & 0.48 & 2.8e3 & \textcolor{green}{\ding{51}}\\
        \midrule
        \multirow{5}{*}{VICUNA} & \multirow{5}{*}{7B} & \multirow{2}{*}{Unstructured} & Magnitude  & 0.03 & 2.29 & 0.06 & 3.40 & 0.11 & 14.31 & \textcolor{green}{\ding{51}}\\
        & && Wanda  & 0.02 & 0.48 & 0.05 & 0.99 & 0.11 & 1.93 & \textcolor{green}{\ding{51}}\\
        \cline{3-11}
        & & \multirow{2}{*}{Semi-structured} & Magnitude  & 0.02 & 1.13 & 0.05 & 2.09 & 0.13 & 15.59 & \textcolor{green}{\ding{51}}\\
        & & & Wanda  & 0.01 & 0.40 & 0.03 & 0.93 & 0.11 & 1.97 & \textcolor{green}{\ding{51}}\\
        \cline{3-11}
        & & Structured & Wanda  & 0.17 & 1.9e2 & 0.17 & 2.4e2 & 0.16 & 4.4e2 & \textcolor{green}{\ding{51}}\\
        \midrule
        \multirow{2}{*}{Mixtral} & \multirow{2}{*}{56B} & Unstructured & Magnitude  & 0.02 & 0.88 & 0.06 & 0.88 & 0.16 & 3.79 & \textcolor{green}{\ding{51}}\\
        \cline{3-11}
        & & Semi-structured & Magnitude  & 0.02 & 0.21 & 0.06 & 0.38 & 0.18 & 0.65 & \textcolor{green}{\ding{51}}\\
        \midrule
        \multirow{2}{*}{LLaVA} & \multirow{2}{*}{7B} & Unstructured & Magnitude  & \multicolumn{6}{c}{L-Min:0.00 L-Max:2.20 V-Min:0.00 V-Max:1.90} & \textcolor{green}{\ding{51}}\\
        \cline{3-11}
        & & Semi-structured & Magnitude  &\multicolumn{6}{c}{L-Min:0.00 L-Max:2.90 V-Min:0.00 V-Max:1.20} & \textcolor{green}{\ding{51}}\\
	\bottomrule
	\end{tabular}

\label{tab:1}
\vspace{-1pt}
\end{table*}

\section{Related Work}

\textbf{Layerwise Sparsity for Pruning.} Early approaches used uniform pruning \cite{lecun1989optimal,hu2016network}, where all layers are pruned at the same sparsity. However, \cite{frankle2018lottery} highlighted that different layers have varying importance, making this strategy suboptimal. To address this issue, some methods \cite{molchanov2019importance,molchanov2016pruning,nonnenmacher2021sosp,zhang2022carrying} automatically determine layerwise sparsity by selecting critical parameters across the entire network. Other studies \cite{he2018amc,yu2021auto} treat pruning as a search problem to identify layerwise sparsity.  In contrast to  these techniques for CNN models in vision tasks, our method focuses specifically on LLMs.\par

\textbf{LLM Pruning.} 
In contrast to traditional techniques, LLM pruning emphasizes data and time efficiency, meaning that pruned models do not require extensive retraining. LLM-Pruner \cite{ma2023llm} offers a structured pruning method based on model dependency relations and uses LoRA to recover the pruned model's performance. SparseGPT \cite{frantar2023sparsegpt} introduces an effective Hessian matrix estimation technique in large-scale models. In addition, Wanda \cite{sun2024pruning} adopts a direct strategy based on the product of weights and activation values to eliminate weights. These methods apply a uniform pruning rate across all layers. Recently, several approaches have been proposed to achieve non-uniform layerwise sparsity. BESA \cite{xu2024besa} employs a differentiable approach to search for optimal layerwise sparsity. 
 DSA \cite{li2024discovering} utilizes a distribution function to allocate sparsity to layers, with the function being searched through evolutionary algorithms. ALS \cite{li2024adaptive} develops an adaptive sparsity allocation strategy based on evaluating the relevance matrix using linear optimization. OWL \cite{yin2024outlier} linearly maps the non-outlier ratio of each layer to its sparsity. These methods rely on search processes or simple linear functions to derive sparsity or allocation strategies. However, for the high-dimensional search space of layerwise sparsity, they cannot guarantee achieving optimal solutions. 
 To address this, we conduct dense empirical research to summarize LLM-specific pruning principles and propose a sparsity allocation strategy that satisfies these principles.\par

\section{Empirical Study}



Traditional approaches typically apply uniform pruning ratios across all layers, limiting overall pruning ability. Methods that allocate layerwise sparsity through heuristics or search strategies also struggle to achieve optimal solutions. To address this, we analyze the sensitivity of individual layers to pruning to understand their importance to overall performance. Based on this analysis, we investigate layerwise sparsity patterns to develop LLM-specific pruning strategies. In this section, we aim to derive the principles that layerwise sparsity in LLMs should follow through  empirical studies.

\begin{table}
    \centering
    \setlength{\tabcolsep}{1mm}
        \caption{Sensitivity of different layers on seven zero-shot tasks. See Appendix A for details and more results.}
    \begin{tabular}{ccccccc}
        \toprule
        \multirow{2}{*}{Model}  & \multicolumn{2}{c}{LLaMA2-7B} &  \multicolumn{2}{c}{OPT-6.7B} & \multicolumn{2}{c}{LLaMA2-13B} \\
        \cline{2-7}
        & 0 & 31 & 0 & 2 & 0 & 39\\
        \midrule
        Wanda     & 0.18 & 2.15 & 0.40 & 12.79 & 0.17 & 0.95 \\

        \midrule
        Magnitude &0.38& 6.03 & 0.42 & 3.63 & 0.83 & 1.92 \\
        \bottomrule
    \end{tabular}

    \label{tab:zero-shot}
\end{table}
\textbf{Layerwise Pruning Sensitivity (LPS).} Our preliminary research primarily focuses on Layerwise Pruning Sensitivity (LPS), which can be utilized to measure the sensitivity of each layer to pruning. Pruning quality is evaluated based on post-pruning performance. Thus, we define pruning sensitivity as the performance degradation after pruning. By measuring the pruning sensitivity of each layer, we can obtain the LPS of the whole model.\par

To better describe our approach, necessary notations are introduced first. Let the original model $M$ have $L$ layers. We define $P(M,\boldsymbol{R}^l)$ as the sparse model obtained by applying pruning to the $l$-th layer of $M$, where $P(\cdot)$ represents the pruning operation. The sparsity ratio $\boldsymbol{R}^l=[0,...,r,...,0]$ specifies that pruning is applied only to $l$-th layer, while the sparsity ratios for all other layers are set to zero. To quantify the pruning sensitivity $S^l$ of the $l$-th layer, we measure the performance difference between the original model and the pruned model. Specifically, we calculate:
$$
S^l=|acc(M)-acc(P(M,\boldsymbol{R}^l))|,
$$
where $acc(\cdot)$ represents the performance evaluation metric (\emph{e.g.}, accuracy). By repeating this process for all layers, we obtain the layerwise pruning sensitivity (LPS), denoted as:
$$\text{LPS}=[S^1, S^2,...,S^L].$$
Based on LPS, we conducted the following empirical study to better understand the role of non-uniform pruning in LLMs.

\begin{figure*}[t]
\vskip -0.10in
\centering

\begin{minipage}[b]{0.49\textwidth}
  \centering
  \includegraphics[width=0.95\textwidth]{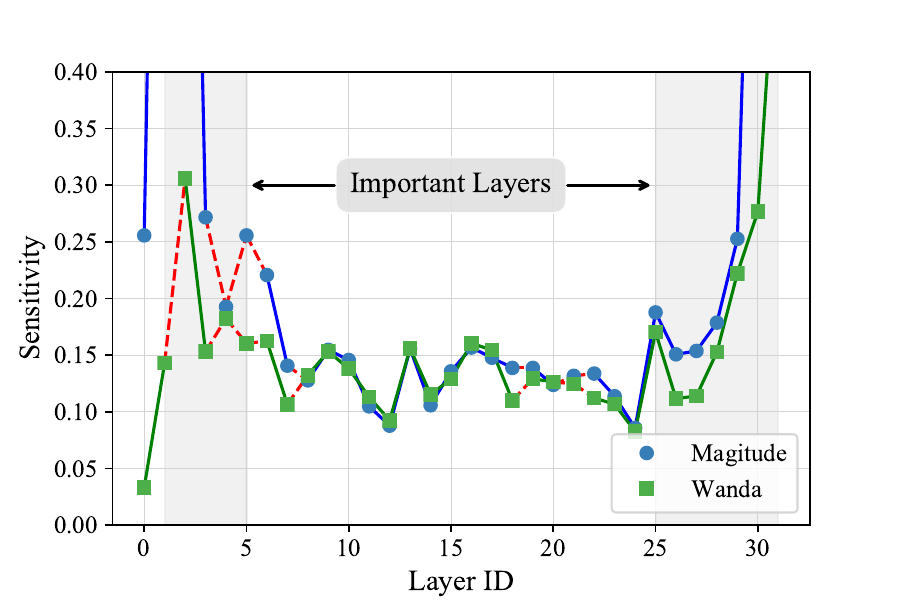}
  \label{sen:1}
  \subcaption{LLaMA2-7B} 
  
\end{minipage}
\hfill
\begin{minipage}[b]{0.49\textwidth}
  \centering
  \includegraphics[width=0.95\textwidth]{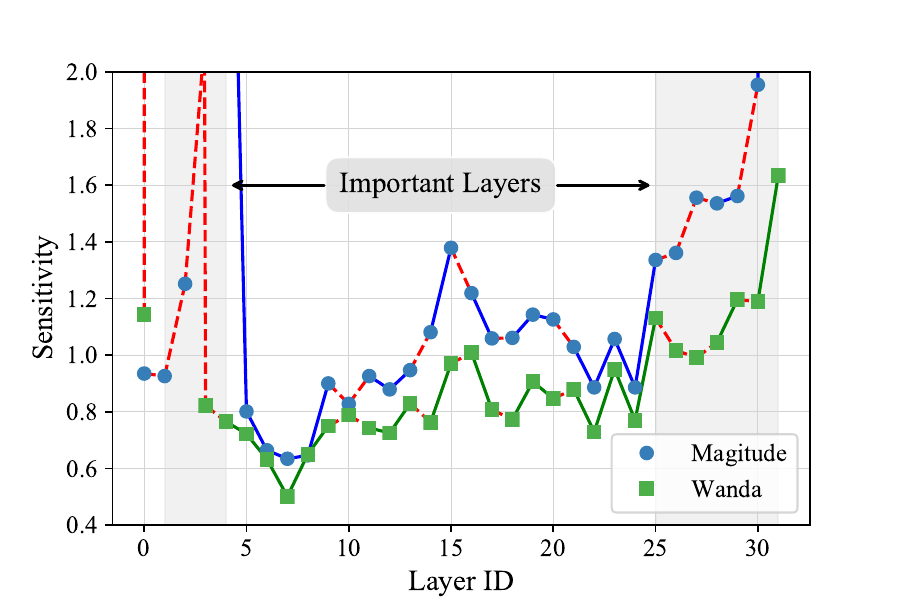}
  \label{sen:2}
  \subcaption{OPT-6.7B} 
  
\end{minipage}
\vspace{-5pt}
\caption{The LPS (increase in WikiText2 perplexity) of the model under different pruning metrics. The red line highlights layers where sensitivity reversal occurs, marking only adjacent layers for clarity. See Appendix A for more results.}
\vskip -0.05in
\label{sen}
\end{figure*}
\subsection{Empirical Study I: LLMs vs. LPS} \label{sec:3.1}

To comprehensively investigate whether pruning LLMs requires differentiated treatment across individual layers, we analyzed the LPS of the model under various settings, including architecture (LLaMA2-(7B, 13B), OPT-6.7B, VICUNA-7B \cite{zheng2023judging}, Mixtral-56B \cite{jiang2024mixtral}, and LLaVA-7B \cite{liu2024visual}), pruning metrics (Magnitude and Wanda), pruning granularity (unstructured, semi-structured, and structured), and tasks (language model and zero-shot tasks). If the LPS of the model exhibits a non-uniform pattern, it implies that pruning LLMs should adopt non-uniform layerwise sparsity ratios, and vice versa.\par
\textbf{Results: The LPS of LLMs exhibits a highly non-uniform pattern across layers.} The complete experimental results are provided in Appendix A, with a summary of the results on the language modeling task (WikiText2) presented in Table \ref{tab:1}. In Table \ref{tab:1}, LPS is considered non-uniform when the maximum sensitivity exceeds twice the minimum sensitivity. Table \ref{tab:1} illustrates that LPS is non-uniform in all settings, with the maximum sensitivity being thousands of times larger than the minimum in some cases. Additionally, this disparity grows progressively as the sparsity level increases. Additionally, we present results on zero-shot tasks in Table \ref{tab:zero-shot}, where the LPS similarly demonstrates non-uniform behavior. The observed non-uniformity reflects the varying importance of layers in LLMs. Therefore, uniform pruning may degrade performance, especially in the case of high sparsity. \emph{Differentiated pruning strategies, which assign different sparsity ratios to layers, are crucial for preserving accuracy while increasing sparsity.}\par

In addition to the standard architectures discussed above, we also conducted experiments on specialized architectures, including Mixtral-56B (an MoE model) and LLaVA-7B (a multimodal model). For Mixtral-56B, the LPS is non-uniform, and the gap between the maximum and minimum sensitivity increases as the sparsity rises. Similarly, for LLaVA-7B, the LPS of the \textbf{V}isual and \textbf{L}anguage heads is also non-uniform. These results provide valuable insights for future pruning efforts on specialized architectures.
\begin{table}
    \centering
    \setlength{\tabcolsep}{1mm}
    \caption{Pruning sensitivity reversal for multiple LLMs.}
    \begin{tabular}{ccccc}
        \toprule
        \multirow{2}{*}{Model}  & LLaMA2 & BaiChuan & OPT & LLaMA2 \\
        & 7B & 7B & 6.7B & 13B\\
        \midrule
        Is there a reversal?     & \textcolor{green}{\ding{51}} & \textcolor{green}{\ding{51}}     & \textcolor{green}{\ding{51}} & \textcolor{green}{\ding{51}} \\

        \midrule
        Reversal rate     & 21.37\% & 23.08\%     & 23.39\%  & 9.62\%\\
        \bottomrule
    \end{tabular}
    
    \label{tab:metric}
    \vspace{-10pt}
\end{table}
\begin{table}
    \centering
    \caption{WikiText2 perplexity with LLaMA2-7B of various metric.}
    \begin{tabular}{cccc}
        \toprule
        Sparsity  & Pruning Metric & Uniform & OWL \\
        \midrule
        \multirow{2}{*}{0.5}     & Magnitude & \textbf{16.03}     & 17.72 \\
     & Wanda & 6.92     & \textbf{6.86} \\

        \midrule
        \multirow{2}{*}{0.7}     & Magnitude & \textbf{49911.45}     & 59240.68 \\
     & Wanda & 80.40     & \textbf{30.22} \\

        \bottomrule
    \end{tabular}

    \label{tab:metirc}
    \vspace{-10pt}
\end{table}

\subsection{Empirical Study II: Pruning Metric vs. LPS}
We further investigate the relationship between pruning metrics and LPS. This study aims to explore whether the optimal layerwise sparsity correlates with the pruning metric, \emph{i.e.}, whether different pruning metrics can share a unified layerwise sparsity. To achieve this, we presented the LPS of different models under different pruning metrics.\par

\textbf{Results: The LPS of the model varies across different pruning metrics.} In Fig. \ref{sen}, we illustrate the LPS of the model under various pruning metrics. A notable phenomenon we observed is \textbf{sensitivity reversal}, \emph{i.e.}, the pruning sensitivity ranking of layers changes under different pruning metrics. For example, under Magnitude pruning, the $i$-th layer is more sensitive than the $j$-th layer, whereas under Wanda pruning, the $j$-th layer becomes more sensitive than the $i$-th layer. This sensitivity reversal is consistently observed across multiple model architectures and is also evident in important layers. Table \ref{tab:metric} provides further statistics on this phenomenon. As shown in the table, the reversal rate is at least around 10\%, and for some low-parameter LLMs, it can exceed 20\%. These results suggest that LPS is metric-dependent. Thus, the layerwise sparsity should be adjusted according to the specific pruning metric. Furthermore, in Table \ref{tab:metirc}, we compare model performance under different layerwise sparsity ratios for each pruning metric. The results reveal that the effectiveness of layerwise sparsity ratios varies with the metric. For example, uniform pruning performs better under Magnitude pruning, while OWL yields superior results under Wanda pruning. \emph{This highlights the need to adjust layerwise pruning ratios based on the chosen metric, as their interaction directly affects model performance}.
\begin{table}
    \centering
    \caption{WikiText2 perplexity and layerwise redundancy differences of different pruning methods.}
    \vspace{-5pt}
    \begin{tabular}{cccc}
        \toprule
        Model  & Method & Perplexity & Max-Min (\%) \\
        \midrule
        \multirow{4}{*}{LLaMA2-7B}     & Global & 5016.12     & 12.32 \\
     & Uniform & 80.4     & 4.42 \\
            & ER & 80.4     & 4.43 \\
             & OWL & 30.47     & 3.43 \\

        \midrule
        \multirow{4}{*}{OPT-6.7B}     & Global & 13792.27     & 70.1 \\
     & Uniform & 162.92     & 33.17 \\
            & ER & 162.81     & 33.19 \\
             & OWL & 40.22     & 21.25 \\

        \bottomrule
    \end{tabular}

    \vspace{-13pt}
    \label{tab:plain}
\end{table}
\subsection{Empirical Study III: Sparse Model vs. Layerwise Redundancy Level}
Section \ref{sec:3.1} demonstrates that pruning some layers to high sparsity does not affect the overall model performance, while pruning others can lead to performance collapse. This suggests that redundancy levels vary across layers in LLMs. As critical information is often stored in weight outliers, the redundancy level can be defined as the non-outlier ratio (NOR). For a layer with input $\mathbf{X} \in \mathbb{R}^{N \times C_{in}}$ and weight $\mathbf{W}\in \mathbb{R}^{C_{out} \times C_{in}}$, its redundancy level is
\begin{equation}
 D=1-\frac{\sum_{i=1}^{C_{\text {out }}} \sum_{j=1}^{C_{in }} \mathbb{I}\left(\mathbf{A}_{ij}>\mathbf{M} \cdot \overline{\mathbf{A}}\right)}{C_{in} C_{out }} 
 \label{eq1}
\end{equation}
where $N$ is the number of tokens, $\mathbf{A}_{i j}=$ $\left\|\mathbf{X}_j\right\|_2 \cdot\left|\mathbf{W}_{i j}\right|$ is the outlier score of weight $\mathbf{W}_{ij}$, $\mathbb{I}(\cdot)$ is the indicator function, and M=5(original settings). OWL aligns layerwise sparsity with layerwise redundancy level, ensuring more uniform redundancy. Inspired by this approach, we hypothesize that the performance of sparse models is correlated with the uniformity of layerwise redundancy level. We validate this hypothesis through experiments using different methods (Global, Uniform, ER \cite{mocanu2018scalable}, and OWL) on LLaMA2-7B and OPT-6.7B. The LRL is calculated based on the C4 \cite{raffel2020exploring} dataset.\par
\textbf{Results: The more uniform the redundancy level across layers in a sparse model, the better its performance.} We quantify the non-uniformity of layerwise redundancy level as the gap between the maximum and minimum redundancy levels (Max-Min). Table \ref{tab:plain} presents the relationship between Max-Min and the performance of sparse models. From the table, it can be concluded that higher non-uniformity in redundancy correlates with lower sparse model performance. The more advanced OWL technique achieves the most uniform layerwise redundancy in the sparse model. The success of OWL demonstrates that \emph{pruning should dynamically adjust layerwise sparsity based on redundancy distribution}, rather than relying on simple uniform sparsity allocation.
\par
\subsection{Empirical Conclusion}
 Based on the results of the three empirical studies discussed above, we summarize three key principles for pruning LLMs: \ding{172} Non-uniform pruning should be applied to LLMs; \ding{173} Layerwise sparsity ratios are dependent on the pruning metric; and \ding{174} The layerwise redundancy level in the sparse model should be as uniform as possible.
\begin{table*}[htbp]
    \centering
    \vskip -0.1in
    \renewcommand{\arraystretch}{0.9}
    \caption{WikiText2 validation perplexity of layerwise sparsity allocation methods at 70\% sparsity.}
    \vspace{-5pt}
    \begin{tabular}{cccccccc}
        \toprule
        \multirow{2}{*}{Metric}  & \multirow{2}{*}{Method} & \multicolumn{3}{c}{LLaMA-2} & LLaMA-3 & OPT  & \multirow{2}{*}{Average}\\
        &&7B & 13B	&70B	&8B	&6.7B & \\
 
        \midrule
        Dense &-& 5.47 & 4.88 & 3.12 & 6.14& 10.13 & 5.95\\
        \midrule
        \multirow{3}{*}{Magnitude} & Uniform & 49911.45 & 214.23 & 423.75 &1625338.50 & 290985.03 & 393374.59 \\
        & OWL	&59240.68	&59.20	&22.21	&1342210.00 	&16547.77	&283615.97\\
        \rowcolor{lightgray!30}
        \cellcolor{white}&MRP & \textbf{14055.67}	&\textbf{58.97}	&\textbf{21.59}	&\textbf{169474.17}	&\textbf{16497.06}	&\textbf{40021.49}\\
        \midrule
        \multirow{3}{*}{Wanda} & Uniform & 80.40&45.42	&10.61	&121.92	&162.92	&84.25 \\
        & OWL	&30.47	&17.91	&9.02	&90.45	&40.22	&37.61\\
        \rowcolor{lightgray!30}
        \cellcolor{white}&MRP & \textbf{23.54}	&\textbf{15.18}	&\textbf{8.57}	&\textbf{63.13}	&\textbf{34.06}	&\textbf{28.90}\\
        \midrule
        \multirow{3}{*}{SparseGPT} & Uniform & 27.52	&19.97	&9.33	&41.84	&\textbf{20.45}	&23.82\\
        & OWL	&20.51	&14.53	&8.20	&36.51	&22.48	&20.45\\
        \rowcolor{lightgray!30}
        \cellcolor{white}&MRP & \textbf{19.19}	&\textbf{12.83}	&\textbf{7.72}	&\textbf{34.99}	&20.83	&\textbf{19.11}\\
        \bottomrule

    \end{tabular}
    \vspace{-0.1in}
    
    \label{tab:lm}
\end{table*}

\section{Maximum Redundancy Pruning}
This section introduces a Maximum Redundancy Pruning (MRP) method and analyses how it satisfies the previously concluded pruning principles.
\begin{table}[t]
    \centering
    \caption{Principles satisfied by the current approach.}
    \vspace{-5pt}
    \begin{tabular}{cccc}
        \toprule
        \multirow{2}{*}{Method}  & Non-uniform & Metric  & Uniform \\
        & sparsity & Dependency & redundancy\\
        \midrule
        Global    & \textcolor{green}{\ding{51}} & \textcolor{green}{\ding{51}}     & \textcolor{red}{\ding{55}} \\
        Uniform    & \textcolor{red}{\ding{55}} & \textcolor{red}{\ding{55}}     & \textcolor{red}{\ding{55}} \\
        ER    & \textcolor{green}{\ding{51}} & \textcolor{red}{\ding{55}}     & \textcolor{red}{\ding{55}} \\
        OWL    & \textcolor{green}{\ding{51}} & \textcolor{red}{\ding{55}}     & \textcolor{green}{\ding{51}} \\
        MRP    & \textcolor{green}{\ding{51}} & \textcolor{green}{\ding{51}}     & \textcolor{green}{\ding{51}} \\

        \bottomrule
    \end{tabular}
    
    \vspace{-10pt}
    \label{tab:5}
\end{table}

\subsection{Method}
The empirical studies above reveal three key principles that need be met when pruning LLMs. However, existing methods do not fully consider these principles (Table \ref{tab:5}). Global pruning often excessively prunes some layers, resulting in highly uneven redundancy across layers, which violates Principle 3. Uniform pruning assigns a fixed sparsity to all layers, thus failing to satisfy any of the principles. ER relies solely on the number of neurons per layer, neglecting Principles 2 and 3. While OWL adjusts layerwise sparsity based on layerwise outlier ratio, it is a metric-independent approach and does not consider the impact of pruning metrics on the sparsity. The neglect of these principles leads to suboptimal performance. Therefore,  there is a need of an ideal layerwise sparsity that satisfy all the principles.\par
To address this issue, we propose a Maximum Redundancy Pruning (MRP) strategy that satisfies all three principles. MRP iteratively prunes in the most redundant layers, thereby making the layerwise redundancy as uniform as possible. Specifically, at each iteration, we calculate the Layerwise Redundancy Level (LRL) based on the C4 dataset, denoted as $\text{LRL} = \left[D^1, D^2, \ldots, D^L\right]$, using Eq. \ref{eq1}. Then we prune at the layers with the highest redundancy level. This iterative process is repeated until the global sparsity reaches the target. Additionally, We perform low-sparsity uniform pruning before iterations. At low sparsity, layers have similar sensitivity, so pre-pruning reduces iterations without affecting results.  Note that we assign a distinct sparsity ratio for each Transformer block instead of each layer, which follows OWL. The complete process are provided in Algorithm \ref{alg:algorithm}. The hyperparameter settings are provided in Appendix B.
 \begin{algorithm}[t]

    \caption{Maximum Redundancy Pruning algorithm}
    \label{alg:algorithm}
    \begin{flushleft}
    \textbf{Input}: Original model $M$, Initial pruning ratio $r$, Target pruning ratio $r_T$, Initial pruning step size $s_0$, Minimum pruning step size $s_{min}$, Decay coefficient $\alpha$\\
    \textbf{Output}: Sparse model $\hat{M}$
    \end{flushleft}
    \begin{algorithmic}[1] 

        \STATE Let $\boldsymbol{R} = [r] \times L$, $s = s_0$.   \hfill \small $\triangleright$ Initialize layerwise sparsity.    
        \STATE $\hat{M} = P(M,\boldsymbol{R})$        \hfill \small $\triangleright$ Pruning
        \STATE $r_c$ = check\_sparsity($\hat{M}$) \hfill \small $\triangleright$ Calculate global sparsity.
        \WHILE{$r_c < r_t$}
        \STATE $\text{LRL}$ = check\_NOR($\hat{M}$)\hfill \small $\triangleright$ Calculate LRL by Eq. \ref{eq1}.
        \STATE $ID$ = ARGMAX($\text{LRL}$) \hfill \small $\triangleright$ Select the most redundant layer.
        \STATE $\boldsymbol{R}[ID] = \boldsymbol{R}[ID] + s$ \hfill \small $\triangleright$ Update layerwise sparsity.
        \STATE $\hat{M} = P(\hat{M},\mathbf{R})$
        \STATE $s = \text{MAX}(s * \alpha,s_{min})$ \hfill \small $\triangleright$ Update pruning step size.
        \STATE $r_c$ = check\_sparsity($\hat{M}$)
        \ENDWHILE
        \STATE \textbf{return} Sparse model $\hat{M}$
    \end{algorithmic}
\end{algorithm}
\begin{table*}[htbp]
    \centering
    \vspace{-10pt}
    \setlength{\tabcolsep}{0.9mm}
    
    \renewcommand{\arraystretch}{0.9} 
    \caption{Accuracies (\%) for 7 zero-shot tasks with 70\% sparsity using LLaMA2-7B and 13B.}
    \begin{tabular}{ccccccccccc}
        \toprule
        LLaMA2  & Metric & Method & BoolQ & RTE & HellaSwag & WinoGrande & ARC-e & ARC-c & OBQA & Average \\ 
        \midrule

        \multirow{10}{*}{7B} 
        & Dense & - & 77.71 & 63.18 & 75.00 & 69.06 & 68.90 & 43.09 & 41.20 & 62.59 \\ \cline{2-11}

        & \multirow{3}{*}{Magnitude} 
        & Uniform & 37.95	&53.07	&25.74	&\textbf{49.49}	&28.24	&24.49	&29.00	&35.42 \\

        &  & OWL & 38.75	&52.35	&28.96	&48.38	&31.82	&25.43	&27.80	&36.21 \\

        \rowcolor{lightgray!30}
        \cellcolor{white} & \cellcolor{white} & MRP & \textbf{38.96}	&\textbf{53.07}	&\textbf{37.49}	&49.25	&\textbf{35.23}	&\textbf{25.94} 	&\textbf{31.20}	&\textbf{38.73}\\ \cline{2-11}

        & \multirow{3}{*}{Wanda} 
        & Uniform & 54.46 & 52.71 & 30.60 & 49.09 & 32.15 & 19.62 & 22.60 & 37.32 \\

        &  & OWL & 61.83 & 52.71 & 37.79 & 55.88 & 41.54 & 23.72 & 29.40 & 43.27 \\

        \rowcolor{lightgray!30}
        \cellcolor{white} & \cellcolor{white} & MRP & \textbf{62.20} & \textbf{52.71} & \textbf{46.64} & \textbf{61.72} & \textbf{45.66} & \textbf{25.60} & \textbf{30.40} & \textbf{46.42} \\ \cline{2-11}

        & \multirow{3}{*}{SparseGPT} 
        & Uniform & 64.59 & 54.87 & 41.10 & 58.48 & 42.00 & 24.74 & 30.60 & 45.20 \\

        &  & OWL & 67.13 & 52.71 & 47.69 & 62.04 & \textbf{45.33} & \textbf{25.94} & 31.40 & 47.46 \\

        \rowcolor{lightgray!30}
        \cellcolor{white} & \cellcolor{white} & MRP & \textbf{68.07} & \textbf{52.71} & \textbf{51.15} & \textbf{63.54} & 45.29 & 25.85 & \textbf{32.00} & \textbf{48.37} \\ 
        \midrule

        \multirow{10}{*}{13B} 
        & Dense & - & 80.55 & 65.34 & 78.25 & 72.06 & 71.84 & 47.78 & 43.00 & 65.55 \\ \cline{2-11}
        
        & \multirow{3}{*}{Magnitude} 
        & Uniform & \textbf{38.72}	&52.71	&29.06	&49.33	&30.81	&22.95	&24.40 	&35.43\\

        &  & OWL & 38.38	&52.71	&42.98	&53.12	&34.34	&25.34	&28.20	&39.30\\

        \rowcolor{lightgray!30}
        \cellcolor{white} & \cellcolor{white} & MRP & 38.56 & \textbf{52.71} & \textbf{47.00} & \textbf{57.14} & \textbf{39.73} & \textbf{29.27} & \textbf{29.00} 
 & \textbf{41.92} \\ \cline{2-11}
        
        & \multirow{3}{*}{Wanda} 
        & Uniform & 62.26 & 52.71 & 31.61 & 51.85 & 35.86 & 19.03 & 26.80 & 40.02 \\

        &  & OWL & 64.92 & 52.71 & 49.68 & 60.93 & 49.92 & 28.58 & \textbf{35.60} & 48.91 \\

        \rowcolor{lightgray!30}
        \cellcolor{white} & \cellcolor{white} & MRP & \textbf{68.20} & \textbf{52.71} & \textbf{53.97} & \textbf{65.19} & \textbf{50.84} & \textbf{29.27} & 35.00 & \textbf{50.74} \\ \cline{2-11}

        & \multirow{3}{*}{SparseGPT} 
        & Uniform & 67.55 & 53.07 & 47.00 & 61.80 & 48.19 & 27.22 & 33.00 & 48.26 \\

        &  & OWL & 68.69 & 54.15 & 54.33 & 66.38 & 50.80 & \textbf{31.48} & \textbf{37.60} & 51.92 \\

        \rowcolor{lightgray!30}
        \cellcolor{white} & \cellcolor{white} & MRP & \textbf{74.40} & \textbf{55.23} & \textbf{56.61} & \textbf{67.80} & \textbf{51.43} & 31.00 & 36.40 & \textbf{53.26} \\ 
        \bottomrule
    \end{tabular}

    \label{tab:zero-shot-experiments}
\end{table*}
\begin{table*}[t]
    
    \renewcommand{\arraystretch}{0.9}
    \centering
    \caption{ End-to-end decode latency speedup of LLaMA2-7B-chat-hf using MRP with the DeepSparse inference engine.}
    \begin{tabular}{ccccccccccc}
        \toprule
    Sparsity & Dense & 10\% & 20\% &30\% & 40\%  &50\% & 60\% &70\% &80\% &90\%\\
    \midrule
    Latency (ms) &390.38	&402.71	&409.20 	&399.37	&373.22	&329.24	&240.99	&222.39	&203.65	&181.37 \\
    Throughput (tokens/sec) &2.56	&2.48	&2.44	&2.50 	&2.68	&3.04	&4.15	&4.50 	&4.91	&5.51\\
    Speedup &1.00×	&0.97×	&0.95×	&0.98×	&1.05×	&1.19×	&1.62×	&1.76×	&1.92×	&2.15×\\
    \bottomrule

    \end{tabular}
    \vspace{-10pt}
    \label{tab:lat}
\end{table*}
\subsection{Analysis}
We analyzed how MRP aligns with principles, as follows. \par
\textbf{Non-uniform sparsity.} In each iteration, MRP prunes only in the most redundant layer, resulting in higher sparsity for redundant layers while maintaining lower sparsity for other layers. Clearly, MRP satisfies this principle.\par

\textbf{Metric-dependent sparsity.} In this algorithm, the layerwise redundancy is recalculated at each iteration, with redundancy quantified using the non-outlier ratio. This design accounts for the impact of previous pruning operations on the non-outlier. Since different pruning metrics affect the non-outlier ratio differently, the layerwise sparsity obtained by MRP varies across different pruning metrics. Therefore, MRP satisfies this principle.\par
\textbf{Uniform layerwise redundancy.} According to Principle 3, a better layerwise pruning sparsity leads to a more uniform LRL, \emph{i.e.}, a smaller $R = \text{max}(\text{LRL}) - \text{min}(\text{LRL}).$
\par
Based on the above, we proceed to analyze the MRP method. We assume that the initial layerwise redundancy for the $b$-th iteration is represented as $\text{LRL}_b=[D^1,...,D^i,...,D^j,...,D^L]$, where $D^i$ and $D^j$ represent the maximum and minimum values, respectively. The degree of non-uniformity is $R_b=D^i - D^j$. After pruning the most redundant layer ($i$-th), the non-outlier ratio of the $i$-th layer decreased $\epsilon$, since pruning tends to remove non-outliers. This results in a new layerwise redundancy $\text{LRL}_{b+1}=[D^1,...,D^i-\epsilon,...,D^j,...,D^L]$, where the pruning step size is small enough that $D^i-\epsilon \geq D^j$. The degree of non-uniformity after pruning is given by:
$$
R_{b+1}=\text{max}(\text{LRL}_{b+1})-\text{min}(\text{LRL}_{b+1}) \leq D^i-D^j=R_b.
$$
Therefore, we conclude that pruning the most redundant layer can lead to a more uniform layerwise redundancy, and MRP iteratively repeats this process to achieve uniformity.

\section{Experiments}
In this section, we evaluate MRP's performance across multiple LLMs, including LLaMA2-(7B/13B/70B) \cite{touvron2023llama}, LLaMA3-8B, and OPT-6.7B \cite{zhang2022opt}. Our evaluation protocol is consistent with prior LLM pruning methods, incorporating assessments of both language modeling and zero-shot capabilities. To demonstrate generalizability, we incorporate MRP directly into three metrics, including Magnitude, Wanda, and SparseGPT. The only distinction between these variants lies in their layerwise sparsity ratios. Additional details about the experimental setup can be found in Appendix B.
\vspace{-10pt}
\subsection{Main Experiments}

\textbf{Language Modeling.} In Table \ref{tab:lm}, we report the perplexity of various pruned LLMs at 70\% sparsity. The results can be summarized as follows: (1) \textbf{MRP, as a general layerwise sparsity method, is effective across various scenarios}. It demonstrates its efficacy in reducing perplexity, regardless of pruning metrics (such as Magnitude, Wanda, and SparseGPT), architectures (including LLaMA2, LLaMA3, and OPT), or model sizes (ranging from 7B, 13B, to 70B). (2) \textbf{The benefits of MRP increase significantly as the model size decreases}. Specifically, as LLaMA2 scales from 70B to 7B, the performance gains from MRP show a clear and monotonic increase. In particular, the performance improvement of MRP with Wanda is 2.04 for LLaMA2-70B, while it reaches 56.86 for LLaMA2-7B. This result aligns with the prior finding that smaller models have more non-uniform LPS.

\textbf{Zero-shot Tasks.} To validate MRP's generalizability, we evaluated the zero-shot ability of various sparse LLMs on different zero-shot downstream tasks with prompting, including BoolQ \cite{clark2019boolq}, RTE \cite{wang2018glue}, HellaSwag \cite{zimmer2023perp}, WinoGrande \cite{sakaguchi2021winogrande}, ARC Easy and Challenge \cite{clark2018think}, and OpenBookQA \cite{mihaylov2018can}, as shown in Table \ref{tab:zero-shot-experiments}. These experiments were conducted using the LLaMA2 at a 70\% sparsity. Overall, MRP achieved the highest accuracy across nearly all settings. This result highlights the potential of MRP for more challenging tasks.

\textbf{Inference Speedup.} We analyzed the speedup achieved by MRP, as shown in Table \ref{tab:lat}. The acceleration corresponds to the end-to-end decoding latency of LLaMA2-7B-chat-hf \cite{touvron2023llama}, in the DeepSparse inference engine \cite{deepsparse2021} on a 32-core Intel Xeon Platinum 8358P CPU. The results indicate that when the global sparsity reaches 70\%, the speedup reaches 1.76×. Notably, as the sparsity increases, the acceleration gain becomes even more significant. For example, at 90\% sparsity, the speedup is approximately 2.15×, providing additional motivation for future efforts targeting extreme sparsity.\par

\begin{table}[t]
    \caption{Time overhead used for computing the layerwise redundancy level (in minutes).}
    \setlength{\tabcolsep}{1mm}
    
    \centering
    \begin{tabular}{ccccc}
        \toprule
        Model  & OPT-6.7B &LLaMA2-7B & LLaMA2-13B & LLaMA2-70B\\
        \midrule
        Total Time & 9.51 & 16.6 & 20.93 & 53.04\\
        \bottomrule

    \end{tabular}
    \vspace{-10pt}
    \label{tab:time}
\end{table}
\begin{table}[t]
    \setlength{\tabcolsep}{1mm}
    \caption{The performance of sparse vision models on ImageNet-1K or MM-Vet.}
    \centering
    \begin{tabular}{ccccc}
        \toprule
        \multirow{2}{*}{Model}&\multirow{2}{*}{Method}  & \multicolumn{3}{c}{Sparsity} \\
        \cline{3-5}
        && 50\% & 60\% & 70\% \\
        \midrule
        \multirow{2}{*}{ConvNeXt-Base} & OWL \textit{w.} Wanda & 82.25 &78.97 & 62.97\\
        & MRP \textit{w.} Wanda & \textbf{82.49} & \textbf{79.98} & \textbf{66.57}\\
        \midrule
        \multirow{2}{*}{DeiT-Base} & OWL \textit{w.} Wanda & 78.08 &70.41 & 44.70\\
        & MRP \textit{w.} Wanda & \textbf{78.24} & \textbf{71.69} & \textbf{49.72}\\
        \midrule
        \multirow{2}{*}{LLaVA-1.5-7B} & OWL \textit{w.} Wanda & 31.20 &25.80 & 14.10\\
        & MRP \textit{w.} Wanda & \textbf{31.50} & \textbf{26.40} & \textbf{15.50}\\
        \bottomrule

    \end{tabular}
    \vspace{-10pt}
    \label{tab:vision}
\end{table}
\textbf{Pruning Efficiency.} Compared to other pruning methods, MRP requires the computation of LRL before pruning. To quantify this additional computational cost, we present the time required for LRL computation in Table \ref{tab:time}. Specifically, we measured the accumulated time spent per iteration on LRL computation using NVIDIA A100 GPUs. The results indicate that the maximum computation time can reach approximately one hours. Although the iterative computation of LRL is time-consuming, it is a one-time process and does not impact the subsequent model inference stages. More importantly, our experimental results demonstrate that the derived layer-wise pruning ratios can be effectively applied to different downstream tasks, ensuring the broad applicability of the computed pruning configurations.\par
\textbf{Vision and Multimodal Model Pruning.} We investigated whether the potential of MRP extends to vision and multimodal models. To this end, we evaluated MRP on vision models (ConvNeXt-Base \cite{liu2022convnet} and DeiT-Base \cite{touvron2021training}) on ImageNet-1K \cite{deng2009imagenet} and the multimodal LLaVA model on MM-Vet benchmarks. All models were pruned without fine-tuning using Wanda, and we compared MRP with OWL under varying sparsity levels (50\%, 60\%, and 70\%). Table \ref{tab:vision} demonstrates that while both methods exhibit comparable performance at moderate sparsity levels, MRP consistently outperforms OWL as the pruning ratio increases—likely owing to its superior preservation of critical parameters. Overall, these results demonstrate that MRP can be effectively extended to vision tasks, CNN architectures, and multimodal models.

\begin{table}[t]
    \centering
    
    \caption{Scratch Training vs. Pruning with fine-tuning. WikiText2 perplexity of ``MRP + Wanda" with LoRA fine-tuning.}
    \setlength{\tabcolsep}{1.5mm}
    \begin{tabular}{ccccc}
        \toprule
        \multirow{2}{*}{Model}  & \multirow{2}{*}{Sparsity} & \multirow{2}{*}{Param} & \multicolumn{2}{c}{Perplexity} \\
        \cline{4-5}
        &  &  & With FT & Without FT\\
        \midrule
        LLaMA2-7B & Dense & 7B & \multicolumn{2}{c}{5.47}\\
        \midrule
        LLaMA2-13B & 0.46 & 7B & \textbf{5.42} & 5.60 \\

        \bottomrule

    \end{tabular}
    
    \label{tab:dense}
\end{table}
\begin{table}[t]
    
    \renewcommand{\arraystretch}{0.95}
    \setlength{\tabcolsep}{1mm}
    
    \centering
    \caption{WikiText2 validation perplexity of layerwise sparsity depending on various metrics.}
    \begin{tabular}{ccccc}
        \toprule
        \multirow{2}{*}{Model}&\multirow{2}{*}{Metric}  & \multicolumn{3}{c}{Layerwise Sparsity} \\
        \cline{3-5}
        && Magnitude & Wanda & SparseGPT \\
        \midrule
        \multirow{3}{*}{LLaMA2-7B} & Magnitude & \textbf{14055.67} &NaN & NaN\\
        & Wanda & 26.84 & \textbf{23.54} & 25.61\\
        & SparseGPT &20.01 & 19.57 & \textbf{19.19}\\
        \midrule
        \multirow{3}{*}{LLaMA2-13B} & Magnitude & \textbf{58.97} &62.19 & 60.85\\
        & Wanda & 16.38 & \textbf{15.18} & 15.21\\
        & SparseGPT &13.90 & 13.12 & \textbf{12.83}\\
        \bottomrule

    \end{tabular}

    \label{tab:metric-ex}
\end{table}

\subsection{Ablation Study}

\textbf{Scratch Training vs. Pruning.} We applied ``MRP + Wanda" to the LLaMA2-13B model and compared the resulting sparse model to the LLaMA2-7B with the same parameter size. The sparse model was fine-tuned using just 30,000 tokens from the C4 training dataset. As shown in Table \ref{tab:dense}, the model pruned by ``MRP + Wanda" achieved performance comparable to LLMs from scratch training and even slightly outperformed it after LoRA fine-tuning \cite{hu2021lora}. This is the first time a sparse model has surpassed an LLM from scratch training with the same architecture and parameter size, offering valuable insights into LLM lightweighting.
\par
\textbf{Metric-dependent Layerwise Sparsity.} To validate Principle 2 (\emph{i.e.}, the layerwise sparsity should depend on the pruning metric), we applied layerwise sparsity ratios derived from different metrics to all pruning metrics and evaluated their performance, as shown in Table \ref{tab:metric-ex}. The results reveal a clear pattern: the diagonal settings, where the layerwise sparsity aligns with the corresponding pruning metric, achieve the best performance. Specifically, for LLaMA2-7B, the sparsity derived from Wanda applied to Wanda pruning achieves the lowest perplexity (23.54), and for LLaMA2-13B, the same is observed (15.18). This demonstrates that layerwise sparsity tailored to a specific pruning metric is indeed optimal for that metric, thereby confirming Principle 2.\par
\textbf{Layerwise Redundancy Level of Sparse Models.} To validate that MRP satisfies Principle 3 (\emph{i.e.}, the layerwise redundancy in a sparse model should be as evenly distributed as possible), we present the layerwise redundancy level of sparse models pruned by different methods, as shown in Appendix A. For clarity, we show the non-uniformity of LRL in Table \ref{tab:non-uniform}. The table shows that on both LLaMA and OPT, the sparse models obtained through MRP exhibit more evenly distributed redundancy across layers, even surpassing the NOR-based method (OWL) in terms of uniformity. This indicates that, compared to other methods, MRP better satisfies Principle 3.\par

\par
\begin{table}[t]
    \centering
    \caption{The degree of non-uniformity of the layerwise redundancy level (LRL) of the model using Wanda pruning.}
    \setlength{\tabcolsep}{1.5mm}
    \begin{tabular}{ccccc}
        \toprule
        \multirow{2}{*}{Model} & \multicolumn{4}{c}{Max-Min (\%)} \\
        \cline{2-5}
        &  Global&  Uniform& OWL & MRP\\
        \midrule
        LLaMA-7B & 12.32 & 4.42 & 3.43&\textbf{2.34}\\
        
        OPT-6.7B & 70.10 & 33.17 & 21.25 & \textbf{4.94} \\

        \bottomrule

    \end{tabular}
    
    \vspace{-5pt}
    \label{tab:non-uniform}
\end{table}
\begin{table}[t]
    \setlength{\tabcolsep}{1mm}
    \caption{WikiText2 validation perplexity of different allocation methods with the Wanda metric for LLaMA1-7B.}
    \centering
    \begin{tabular}{ccccc}
        \toprule
        Ratio  & Uniform & BESA & DSA & MRP \\
        \midrule
        65\%&20.85&18.52&12.88 &\textbf{12.22} \\
        
        70\% & 81.18 & 42.58 & 24.50 & \textbf{23.22}\\
        \bottomrule

    \end{tabular}
    
    \label{tab:DSA}
\end{table}
\begin{table*}[t]
    \centering
    \vspace{-5pt}
    \setlength{\tabcolsep}{1.5mm}
    \caption{WikiText2 validation perplexity of various layerwise sparsity.}
    \begin{tabular}{cccccccccccccc}
        \toprule
        \multirow{2}{*}{Metric}  & \multirow{2}{*}{Method} & \multicolumn{6}{c}{LLaMA2-7B} & \multicolumn{6}{c}{LLaMA2-13B} \\
        &&20\% & 30\% & 40\% &50\% & 60\% &70\% & 20\% & 30\% & 40\% &50\% & 60\% &70\% \\
 
        \midrule
        \multirow{3}{*}{Magnitude} & Uniform  &5.71	&6.23	&7.92	&16.03	&1925.01	&49911.45 &4.97	&5.15	&5.64	&6.83	&11.82	&214.23\\
        & OWL	&5.78	&6.48	&8.63	&17.72	&386.23	&59240.68 &4.97	&5.17	&5.66	&6.87	&10.54	&59.20\\
        \rowcolor{lightgray!30}
        \cellcolor{white}&MRP & \textbf{5.70}	&\textbf{6.20}	&\textbf{7.85}	&\textbf{15.99}	&\textbf{337.94}	&\textbf{25815.23} &\textbf{4.96}	&\textbf{5.13}	&\textbf{5.59}	&\textbf{6.66}	&\textbf{9.62}&\textbf{58.97}\\
 
        \midrule
        \multirow{3}{*}{Wanda} & Uniform  &5.59	&5.74	&6.06	&6.92	&10.78	&80.40
&4.99	&5.13	&5.37	&5.97	&8.40	&45.42\\
        & OWL	&5.59	&5.76	&6.10	&6.86	&9.19	&30.47&5.01	&5.14	&5.38	&5.93	&7.50 	&17.91\\
        \rowcolor{lightgray!30}
        \cellcolor{white}&MRP & \textbf{5.58}	&\textbf{5.73}	&\textbf{6.04}	&\textbf{6.81}	&\textbf{9.13}	&\textbf{23.54}&\textbf{4.99}	&\textbf{5.12}	&\textbf{5.36}	&\textbf{5.88}	&\textbf{7.33}	&\textbf{15.18}\\
        \midrule

        \multirow{3}{*}{SparseGPT} & Uniform  &5.61	&5.78	&6.11	&7.00	&10.22	&27.52
&4.98	&\textbf{5.10}	&5.38	&6.03	&8.28	&19.97\\
        & OWL	&5.62	&5.80	&6.16	&6.93	&9.21	&20.51
&4.99	&5.12	&5.40	&6.02	&7.67	&14.53\\
        \rowcolor{lightgray!30}
        \cellcolor{white}&MRP & \textbf{5.61}	&\textbf{5.77}	&\textbf{6.10} 	&\textbf{6.90} 	&\textbf{9.07}	&\textbf{19.19}
&\textbf{4.98}	&5.11	&\textbf{5.37}	&\textbf{5.98}	&\textbf{7.48}	&\textbf{12.83}\\
        \bottomrule

    \end{tabular}
    
    \label{tab:lg}
\end{table*}
\textbf{Comparison More Sparsity Allocation Methods.} We compared the performance of MRP with other sparsity allocation methods on WikiText2 dataset. The results are shown in Table \ref{tab:DSA}. Across all settings, MRP consistently achieved superior performance compared to others. This demonstrates that our method not only surpasses uniform pruning but also achieves superior performance compared to state-of-the-art sparsity allocation methods.\par
\textbf{More Sparsity.} We provide results for more global sparsity in Table \ref{tab:lg}. The results show that MRP outperforms the other two methods at almost all sparsity, demonstrating its generalizability across different sparsity configurations. Furthermore, we observe that as the sparsity decreases, the performance gap between non-uniform pruning methods and uniform pruning methods gradually narrows. Notably, at 20\% sparsity, the performance of all three methods becomes almost identical. This suggests that, at extremely low sparsity, all layers exhibit similar sensitivity to pruning, consistent with previous findings.
\subsection{Effectiveness Analysis}
We study several aspects of MRP to better understand its effectiveness in layerwise sparsity allocation. Current sparsity allocation methods first quantify the redundancy level of each layer, and subsequently perform a one-shot mapping from these redundancy metrics to layer-wise sparsity. Compared with them, our approach takes into account the following two aspects:\par
\begin{figure}
\centering
    \vspace{-10pt}
  \includegraphics[width=0.49\textwidth]{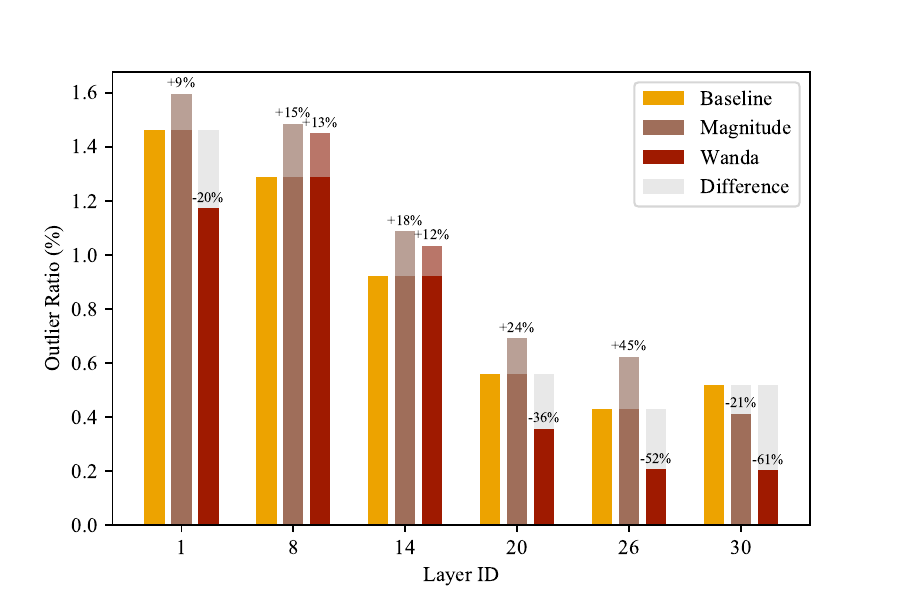}
\caption{Effect of pruning previous layers on layerwise outlier ratio in LLaMA2-7B.}
\vspace{-10pt}
\label{fig:or}
\end{figure}
\textbf{Inter-layer Dependencies.} In existing model architectures, the layers are closely interconnected, with each layer building on the outputs of the previous one. Consequently, pruning a single layer can change the data distribution entering subsequent layers. Since most redundancy metrics depend on the input to each layer, pruning one layer inevitably alters the inputs to downstream layers, thereby changing their redundancy levels, as shown in Fig. \ref{fig:or}. This figure demonstrates that pruning previous layers affects the outlier distribution in subsequent layers, leading to changes in redundancy levels. We also observe that the deepest layers exhibit the most significant changes, as they accumulate the effects of all preceding layer modifications. However, existing methods often map all layers’ redundancy metrics into layer-wise sparsity in a single step. As a result, this approach cannot adapt to changes in downstream layers’ redundancy caused by pruning upstream layers, leading to imbalanced or excessive pruning in some layers. In contrast, our approach prunes only one layer at a time and recalculates redundancy after each pruning phase. By considering the effect of previously pruned layers on subsequent ones, this iterative strategy helps preserve overall model accuracy.\par

\textbf{Allocation Functions.} Existing methods typically map redundancy metrics to sparsity using an allocation function. This function relies on specific assumptions about its functional form, such as linear functions. However, these assumptions may be overly simplistic. As a result, the allocation function might fail to accurately capture the potentially complex relationship between redundancy and sparsity. In contrast, our approach iteratively prunes the layer with the highest redundancy. This strategy ensures that each pruning step is optimal with respect to the network's current configuration. Consequently, it provides a more accurate approximation of the actual relationship between redundancy and sparsity.

\section{Conclusion}
In this paper, we focus a crucial issue in LLM pruning: layerwise sparsity ratios. To enhance the understanding of LLM pruning, we conducted extensive empirical research based on Layerwise Pruning Sensitivity (LPS), leading to the formulation of three principles regarding layerwise sparsity in LLMs: (1) non-uniformity, (2) dependence on pruning metrics, and (3) making the layerwise redundancy level in the sparse model as uniform as possible. Based on these principles, we propose the Maximum Redundancy Pruning (MRP) method, which iteratively prunes the most redundant layers (those with the highest non-outlier ratio), ensuring that the resulting layerwise sparsity adheres to the abovementioned principles. MRP demonstrates promising results on the LLaMA and OPT models. The pruning principles derived in this work are based on extensive empirical studies on various LLMs, demonstrating broad applicability and providing valuable guidance for future pruning efforts in LLMs.


\bibliographystyle{named}
\bibliography{ijcai25}



\clearpage
\appendix
\section{More results}
Due to space limitations, we include part of the empirical study and a selection of experimental results in this section.\par
\subsection{Supplementary Results for Empirical Study 1}

\begin{figure*}
\vskip -0.1in
\centering

\begin{subfigure}[b]{0.33\textwidth}
  \centering
  \includegraphics[width=1\textwidth]{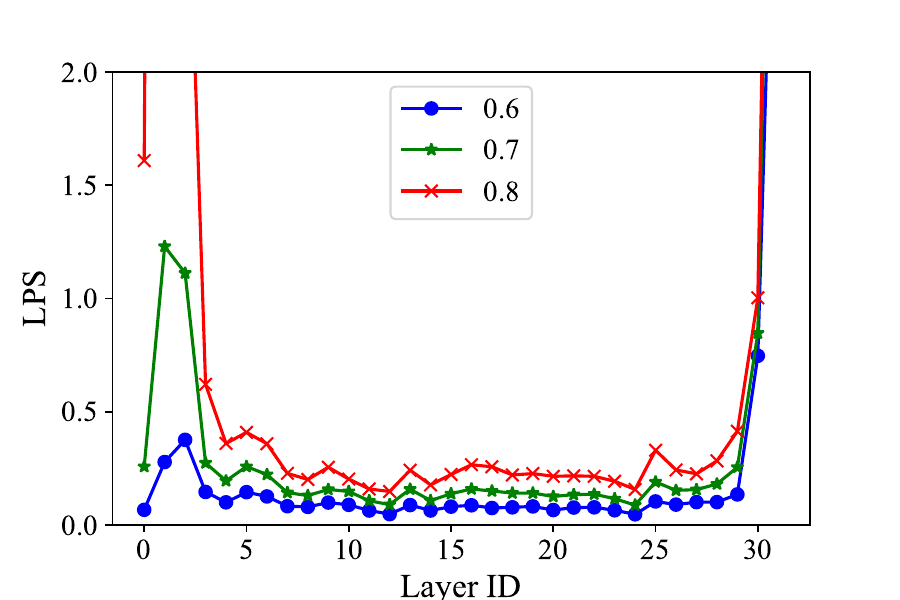}
  \label{OD:1}
  \subcaption{LLaMA2-7B-Magnitude-Unstructured} 
  
\end{subfigure}
\hfill
\begin{subfigure}[b]{0.33\textwidth}
  \centering
  \includegraphics[width=1\textwidth]{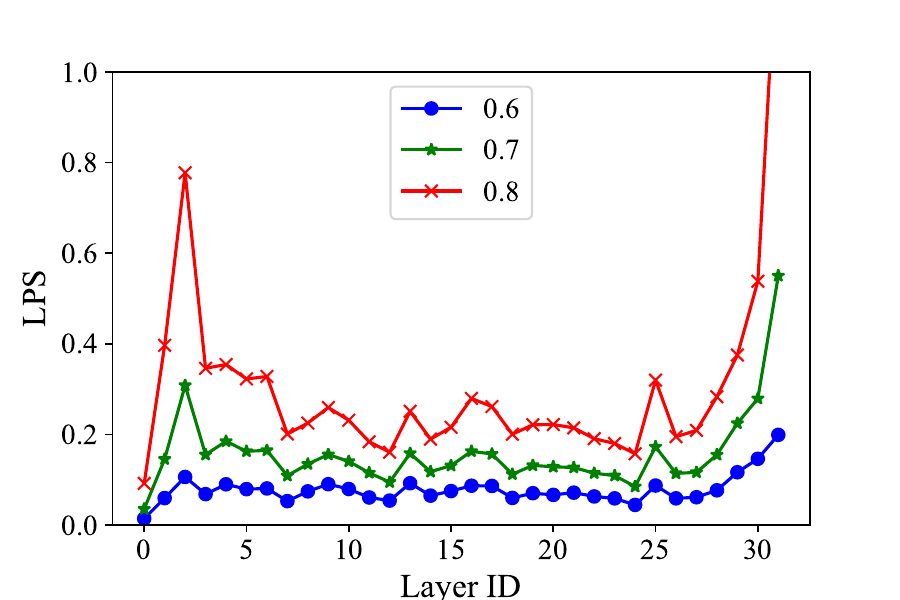}
  \label{OD:2}
  
  \subcaption{LLaMA2-7B-Wanda-Unstructured} 
  
\end{subfigure}
\hfill
\begin{subfigure}[b]{0.33\textwidth}
  \centering
  \includegraphics[width=1\textwidth]{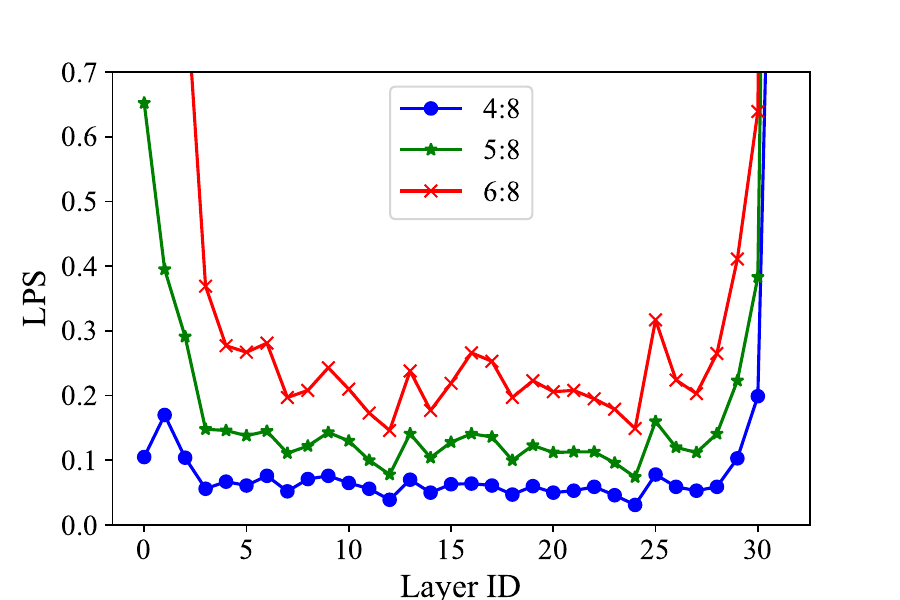}
  \label{OD:2}
  \subcaption{LLaMA2-7B-Magnitude-N:M} 
  
\end{subfigure}
\hfill
\begin{subfigure}[b]{0.33\textwidth}
  \centering
  \includegraphics[width=1\textwidth]{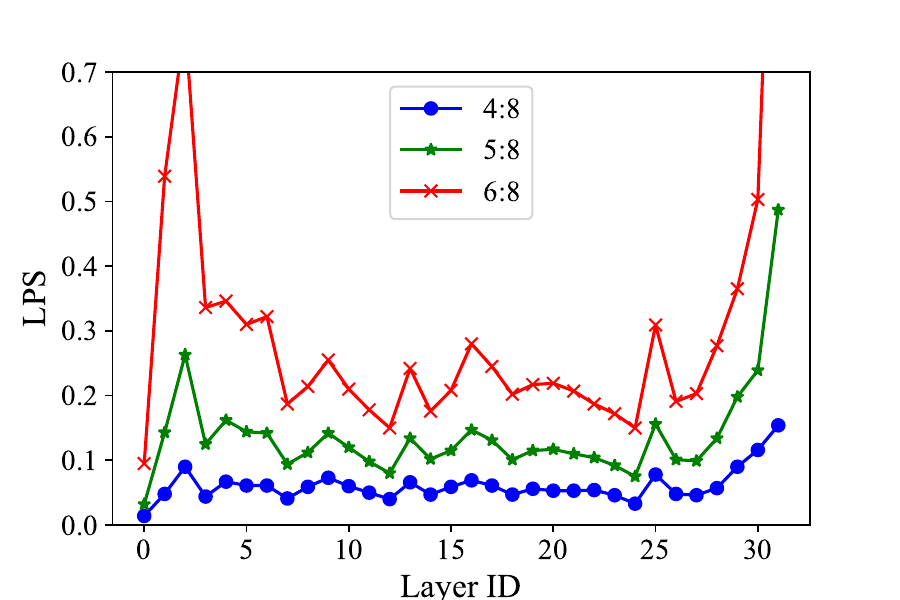}
  \label{OD:2}
  \subcaption{LLaMA2-7B-Wanda-N:M} 
  
\end{subfigure}
\hfill
\begin{subfigure}[b]{0.33\textwidth}
  \centering
  \includegraphics[width=1\textwidth]{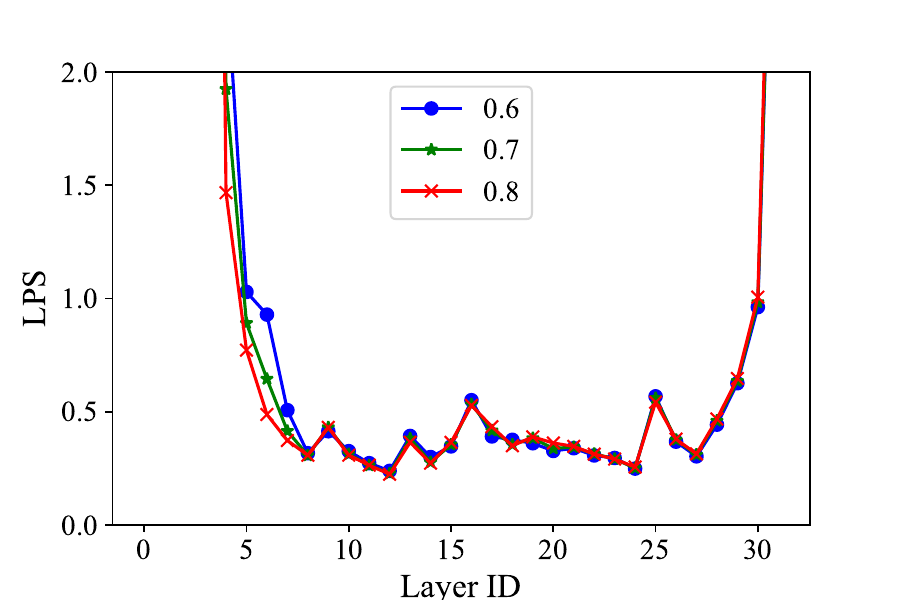}
  \label{OD:2}
  \subcaption{LLaMA2-7B-Wanda-Structured} 
  
\end{subfigure}
\hfill
\begin{subfigure}[b]{0.33\textwidth}
  \centering
  \includegraphics[width=1\textwidth]{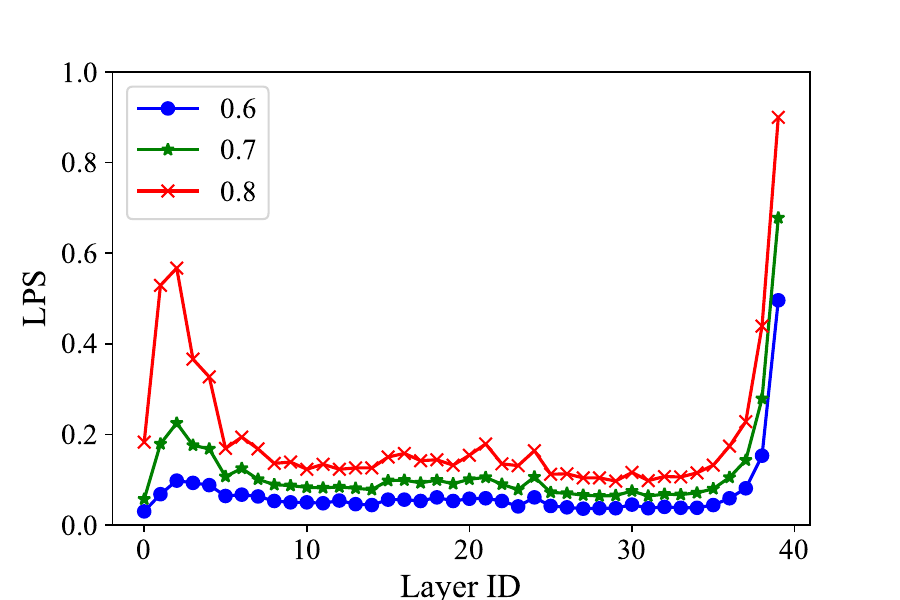}
  \label{OD:1}
  \subcaption{LLaMA2-13B-Magnitude-Unstructured} 
  
\end{subfigure}
\hfill
\begin{subfigure}[b]{0.33\textwidth}
  \centering
  \includegraphics[width=1\textwidth]{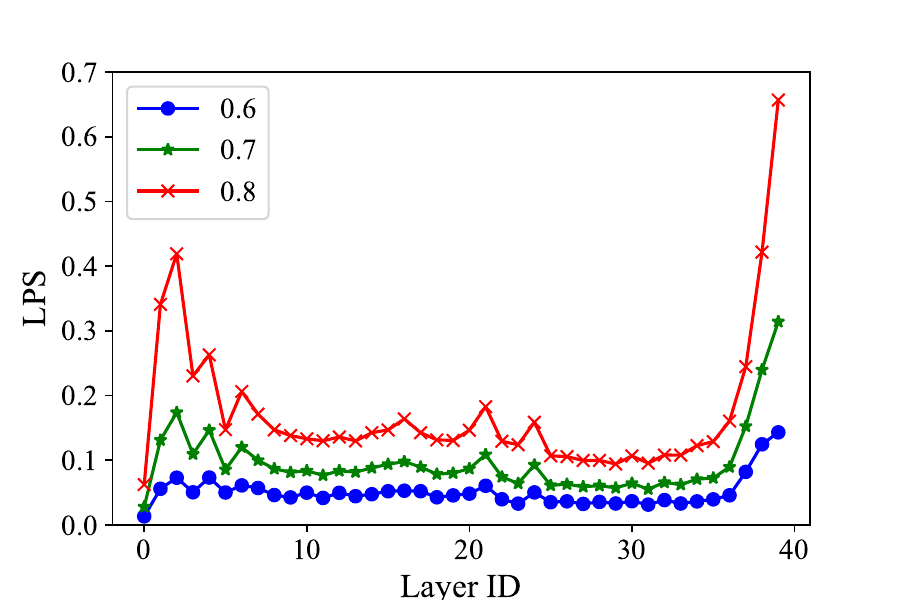}
  \label{OD:2}
  \subcaption{LLaMA2-13B-Wanda-Unstructured} 
  
\end{subfigure}
\hfill
\begin{subfigure}[b]{0.33\textwidth}
  \centering
  \includegraphics[width=1\textwidth]{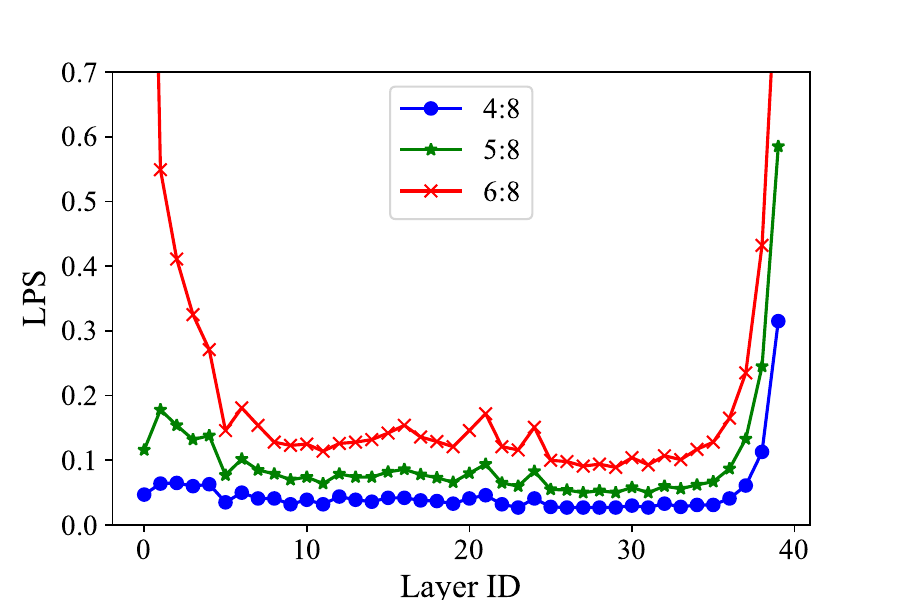}
  \label{OD:2}
  \subcaption{LLaMA2-13B-Magnitude-N:M} 
  
\end{subfigure}
\hfill
\begin{subfigure}[b]{0.33\textwidth}
  \centering
  \includegraphics[width=1\textwidth]{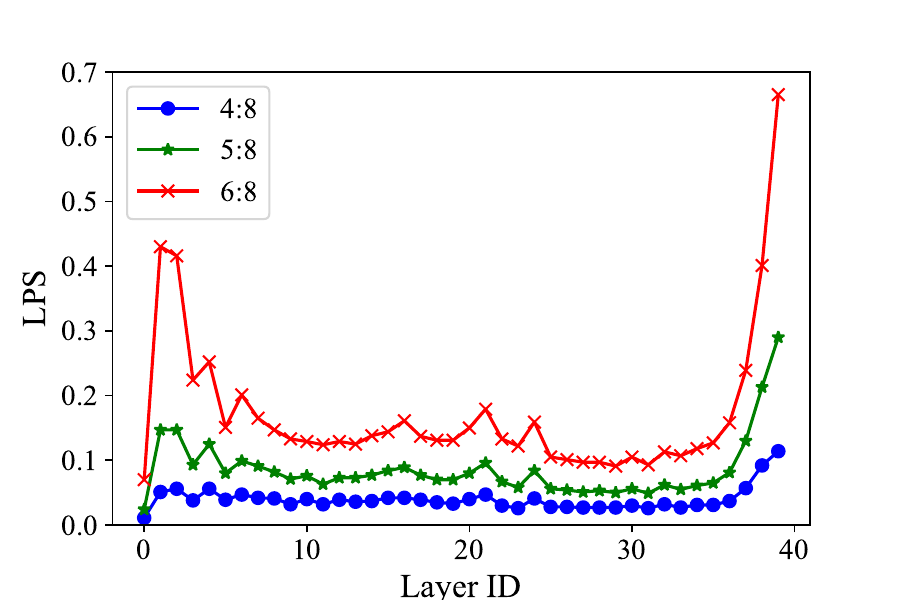}
  \label{OD:2}
  \subcaption{LLaMA2-13B-Wanda-N:M} 
  
\end{subfigure}
\hfill
\begin{subfigure}[b]{0.33\textwidth}
  \centering
  \includegraphics[width=1\textwidth]{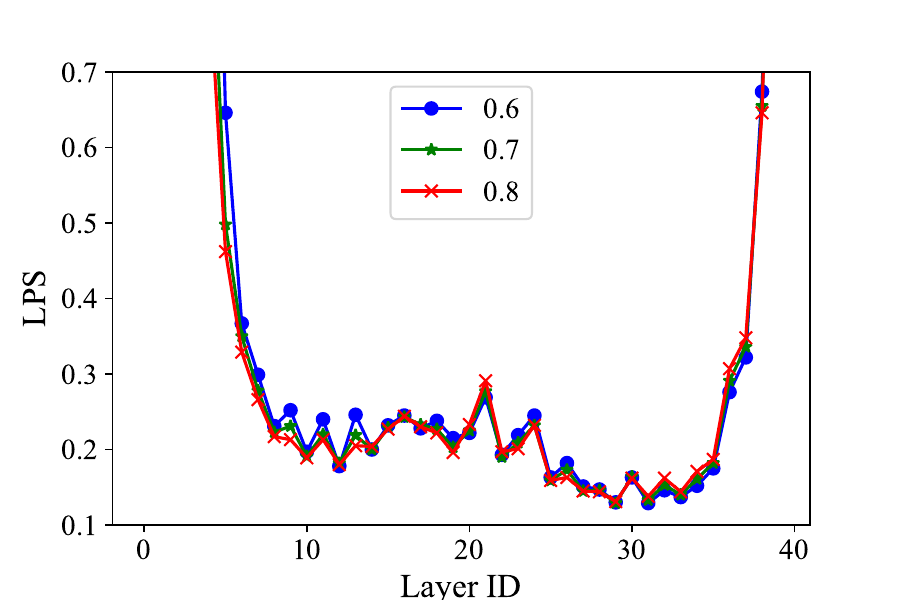}
  \label{OD:2}
  \subcaption{LLaMA2-13B-Wanda-Structured} 
  
\end{subfigure}
\caption{The Layerwise Pruning Sensitivity (LPS) of LLaMA2 at various layerwise sparsity. The results are based on WikiText2 perplexity.}
\vspace{-10pt}
\label{fig1}
\end{figure*}

\begin{figure*}
\vskip -0.1in
\centering

\begin{subfigure}[b]{0.33\textwidth}
  \centering
  \includegraphics[width=1\textwidth]{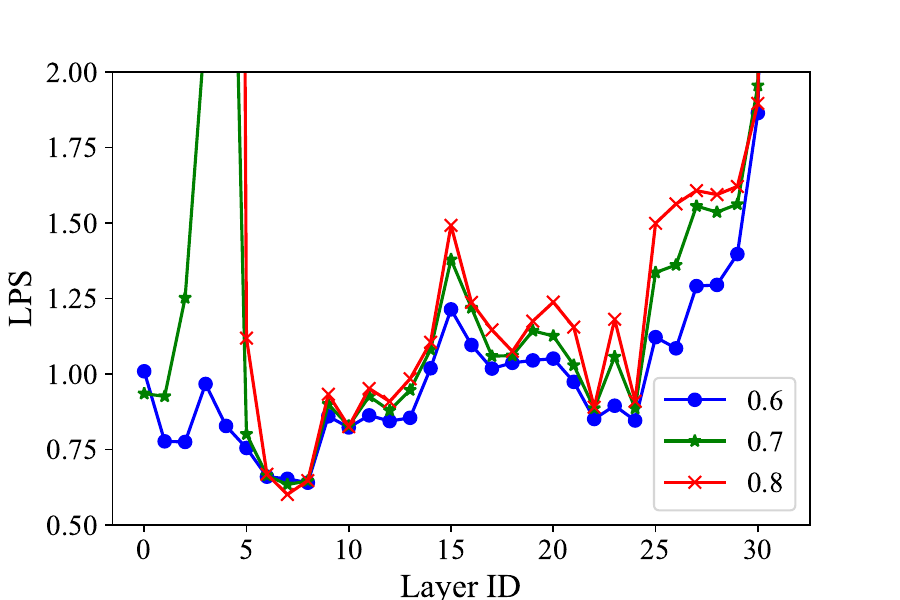}
  \label{OD:1}
  \subcaption{OPT-6.7B-Magnitude-Unstructured} 
  
\end{subfigure}
\hfill
\begin{subfigure}[b]{0.33\textwidth}
  \centering
  \includegraphics[width=1\textwidth]{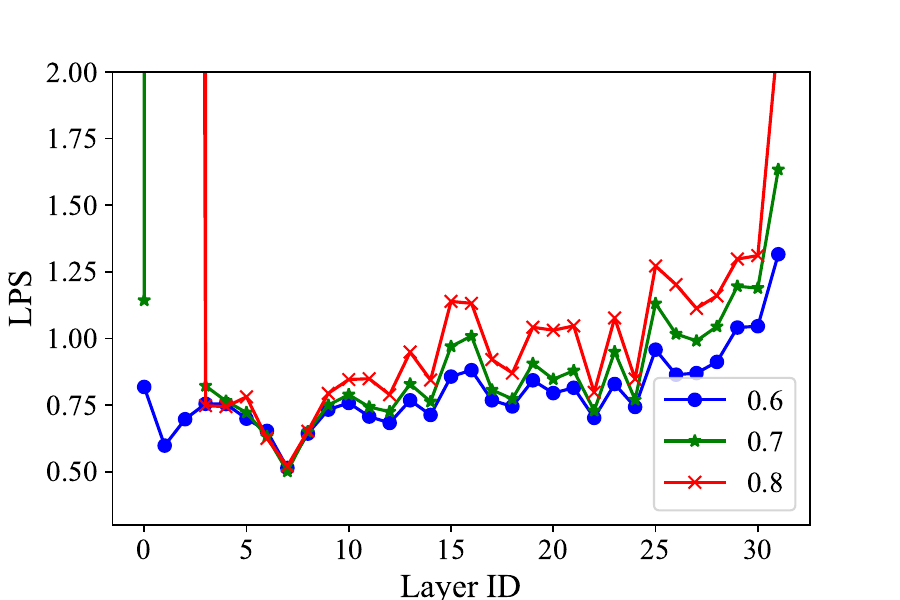}
  \label{OD:2}
  
  \subcaption{OPT-6.7B-Wanda-Unstructured} 
  
\end{subfigure}
\hfill
\begin{subfigure}[b]{0.33\textwidth}
  \centering
  \includegraphics[width=1\textwidth]{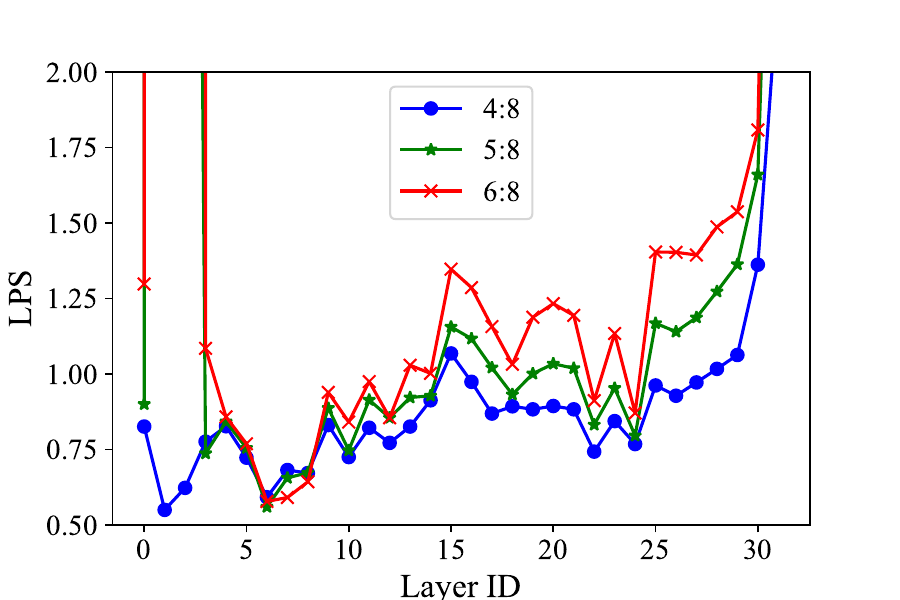}
  \label{OD:2}
  \subcaption{OPT-6.7B-Magnitude-N:M} 
  
\end{subfigure}
\hfill
\begin{subfigure}[b]{0.33\textwidth}
  \centering
  \includegraphics[width=1\textwidth]{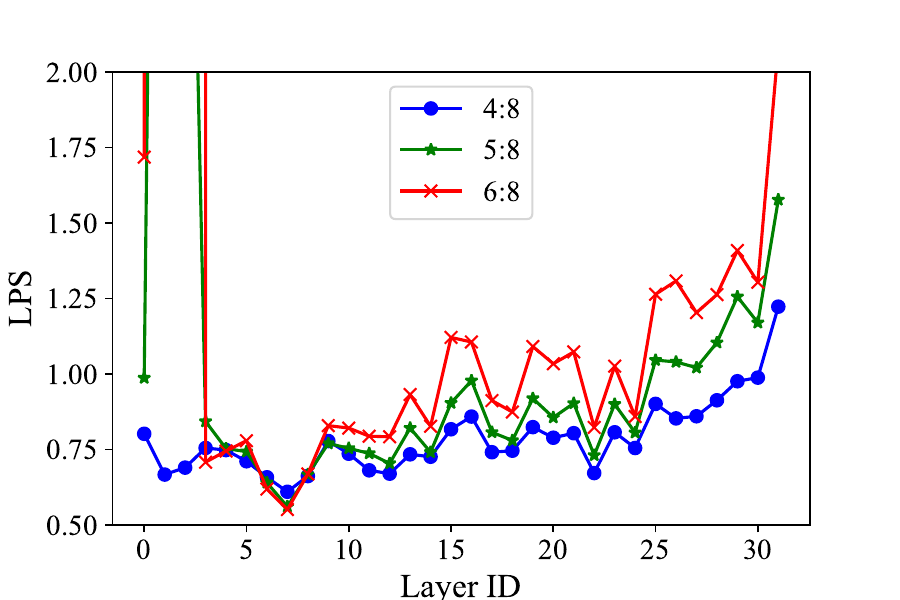}
  \label{OD:2}
  \subcaption{OPT-6.7B-Wanda-N:M} 
  
\end{subfigure}
\hfill
\begin{subfigure}[b]{0.33\textwidth}
  \centering
  \includegraphics[width=1\textwidth]{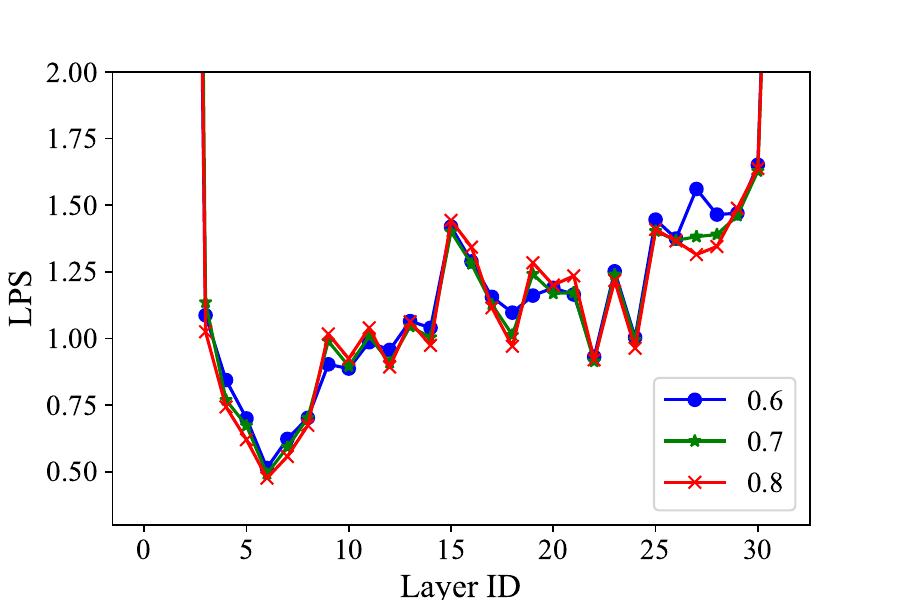}
  \label{OD:2}
  \subcaption{OPT-6.7B-Wanda-Structured} 
  
\end{subfigure}
\hfill
\begin{subfigure}[b]{0.33\textwidth}
  \centering
  \includegraphics[width=1\textwidth]{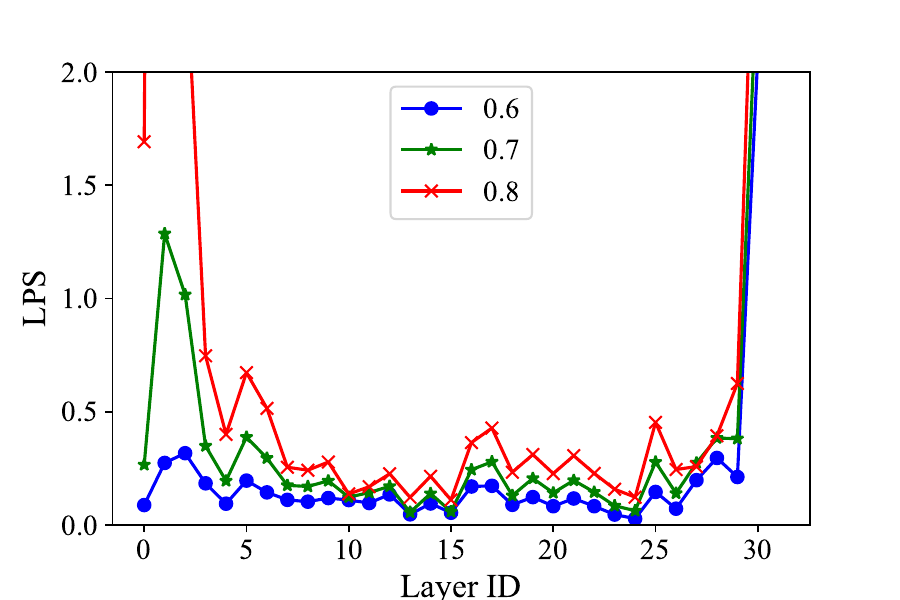}
  \label{OD:1}
  \subcaption{VICUNA-7B-Magnitude-Unstructured} 
  
\end{subfigure}
\hfill
\begin{subfigure}[b]{0.33\textwidth}
  \centering
  \includegraphics[width=1\textwidth]{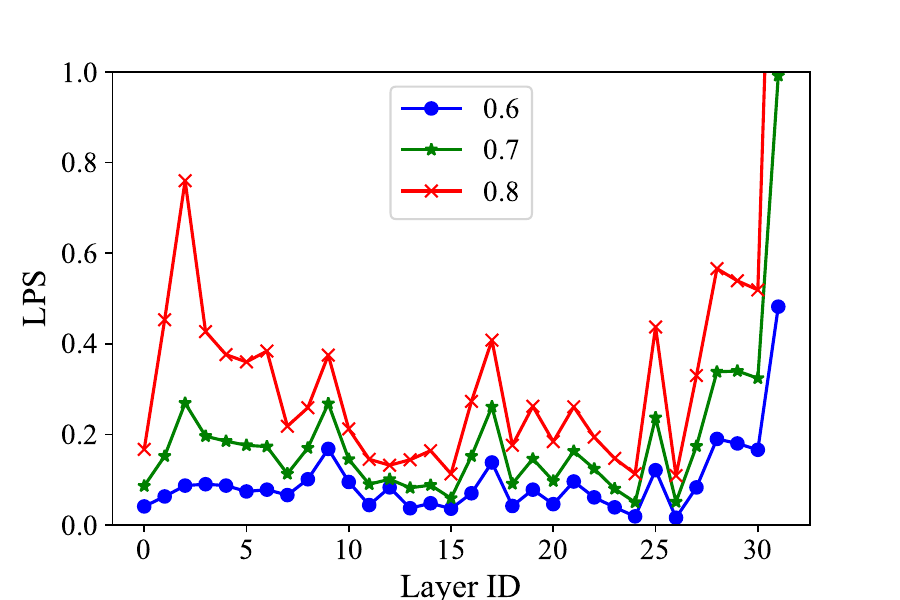}
  \label{OD:2}
  \subcaption{VICUNA-7B-Wanda-Unstructured} 
  
\end{subfigure}
\hfill
\begin{subfigure}[b]{0.33\textwidth}
  \centering
  \includegraphics[width=1\textwidth]{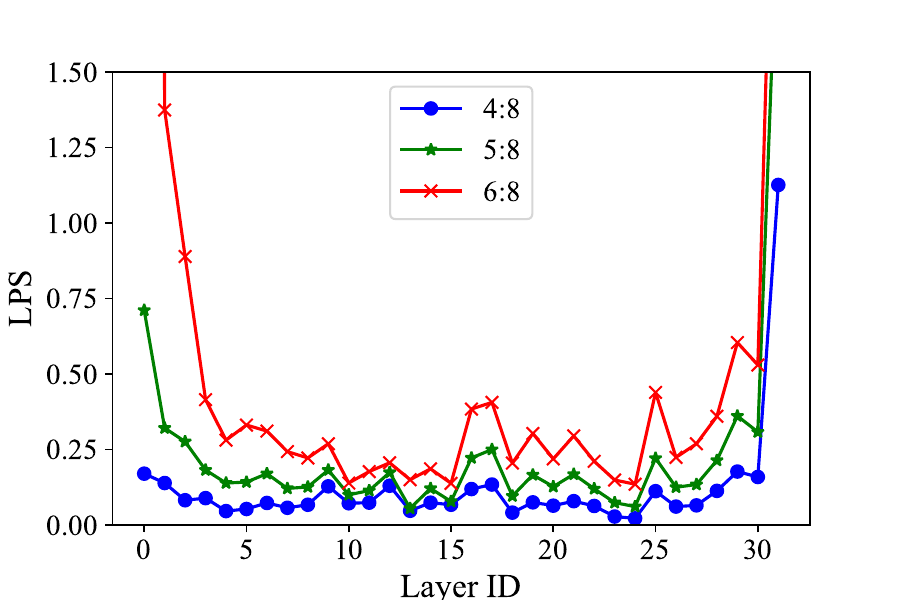}
  \label{OD:2}
  \subcaption{VICUNA-7B-Magnitude-N:M} 
  
\end{subfigure}
\hfill
\begin{subfigure}[b]{0.33\textwidth}
  \centering
  \includegraphics[width=1\textwidth]{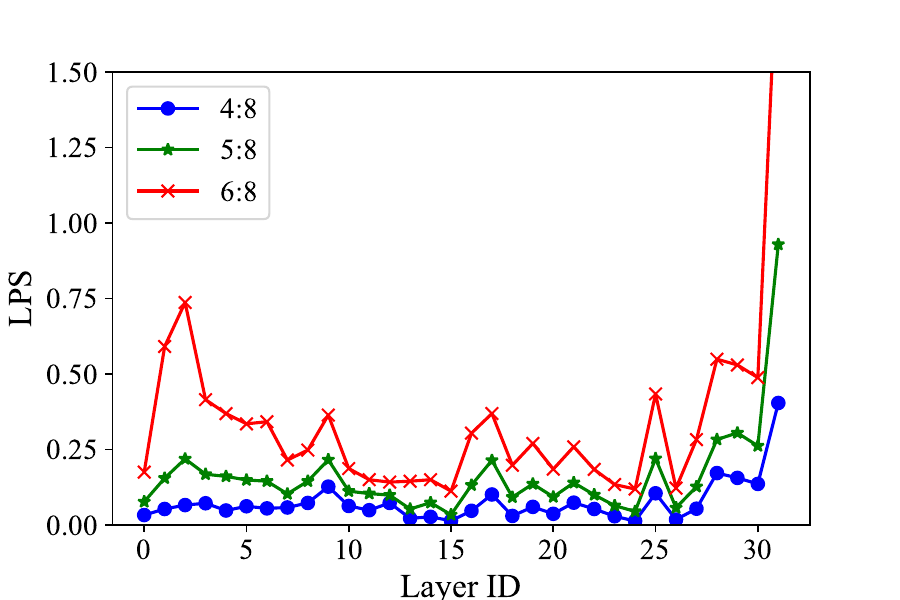}
  \label{OD:2}
  \subcaption{VICUNA-7B-Wanda-N:M} 
  
\end{subfigure}
\hfill
\begin{subfigure}[b]{0.33\textwidth}
  \centering
  \includegraphics[width=1\textwidth]{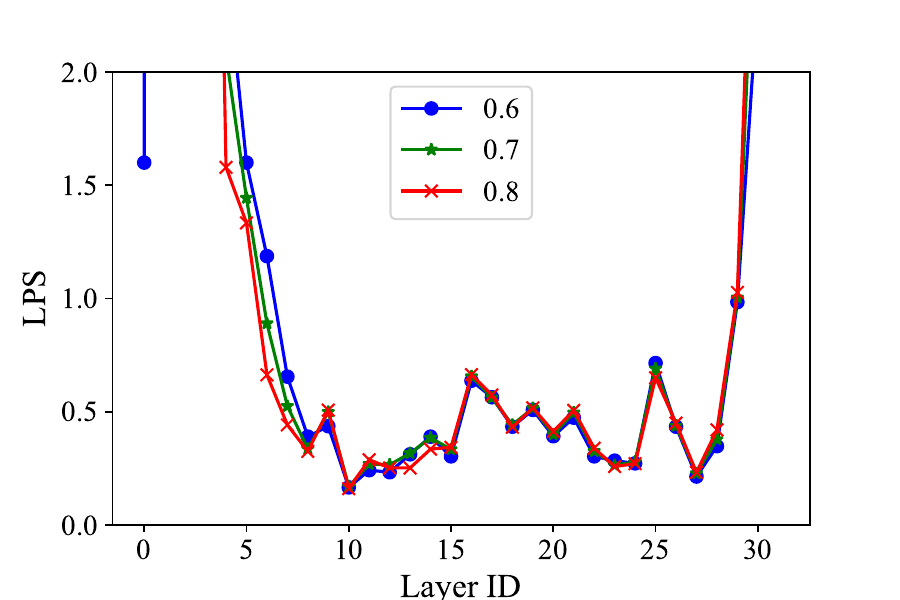}
  \label{OD:2}
  \subcaption{VICUNA-7B-Wanda-Structured} 
  
\end{subfigure}
\caption{The Layerwise Pruning Sensitivity (LPS) of OPT-6.7B and VICUNA-7B at various layerwise sparsity. The results are based on WikiText2 perplexity.}
\vspace{-10pt}
\label{fig2}
\end{figure*}

Fig. \ref{fig1} and \ref{fig2} illustrate the layer-wise pruning sensitivity of multiple LLMs under different settings. From the figures, the following observations can be made: (1) The layer-wise pruning sensitivity of dense LLMs loosely follows a "U" shape, with significant sensitivity at both ends and a monotonic decline in the middle region. (2) Higher sparsity rates result in less uniform layer-wise pruning sensitivity. (3) Finer pruning granularity leads to lower pruning sensitivity.\par
\begin{figure*}
\vskip -0.1in
\centering

\begin{subfigure}[b]{0.33\textwidth}
  \centering
  \includegraphics[width=1\textwidth]{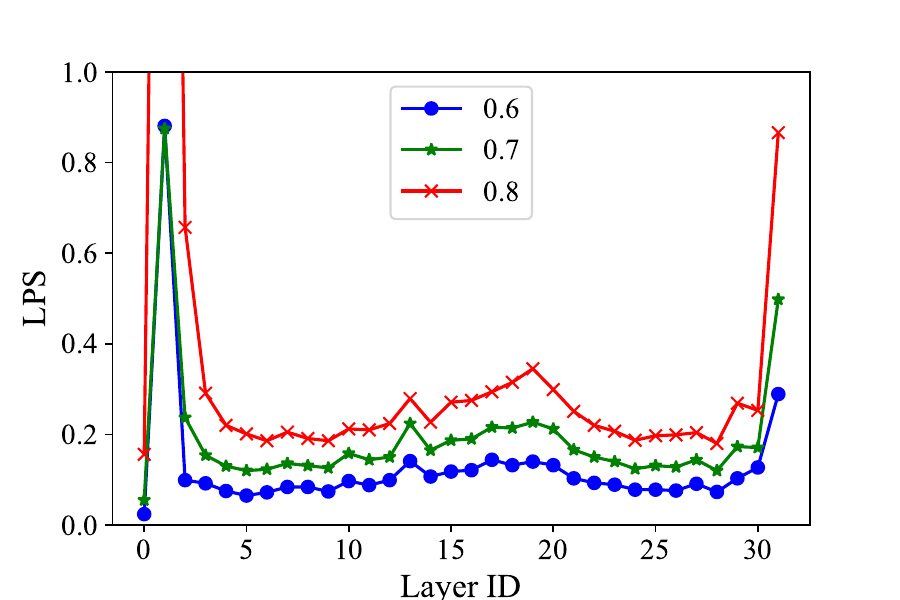}
  \label{OD:1}
  \subcaption{Mixtral-56B-Magnitude-Unstructured} 
  
\end{subfigure}
\hfill
\begin{subfigure}[b]{0.33\textwidth}
  \centering
  \includegraphics[width=1\textwidth]{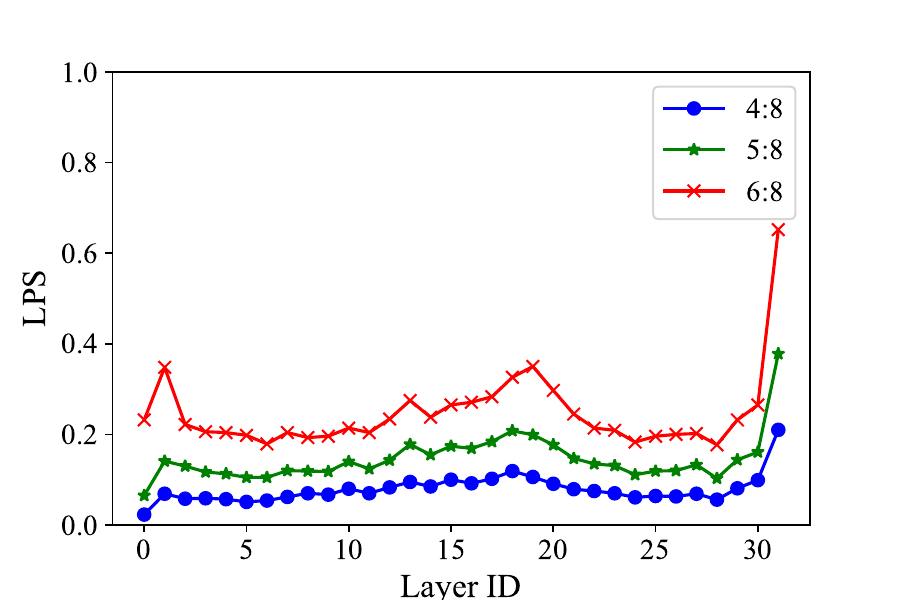}
  \label{OD:2}
  
  \subcaption{Mixtral-56B-Magnitude-N:M} 
  
\end{subfigure}
\hfill
\begin{subfigure}[b]{0.33\textwidth}
  \centering
  \includegraphics[width=1\textwidth]{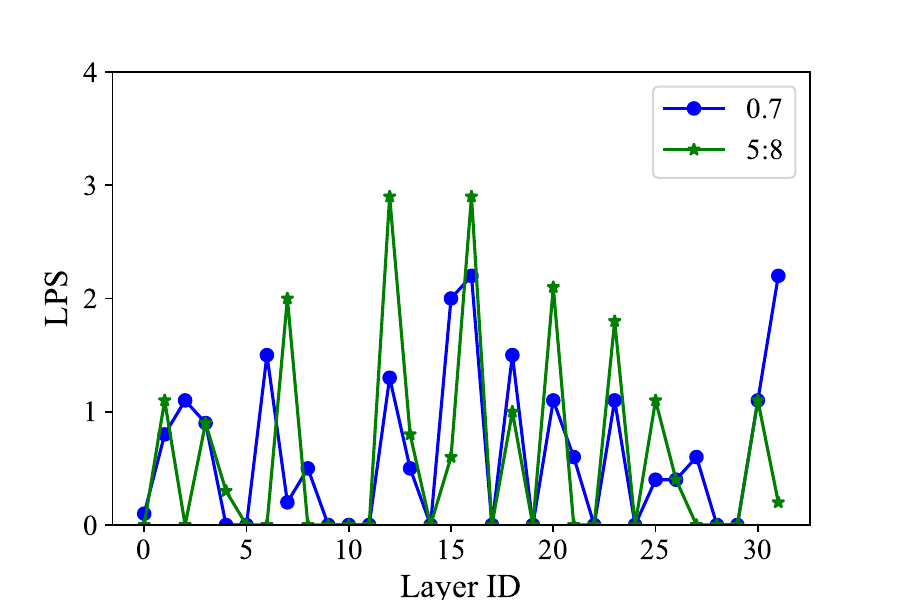}
  \label{OD:1}
  \subcaption{LLaVA-7B-Magnitude-Language} 
  
\end{subfigure}
\hfill
\begin{subfigure}[b]{0.33\textwidth}
  \centering
  \includegraphics[width=1\textwidth]{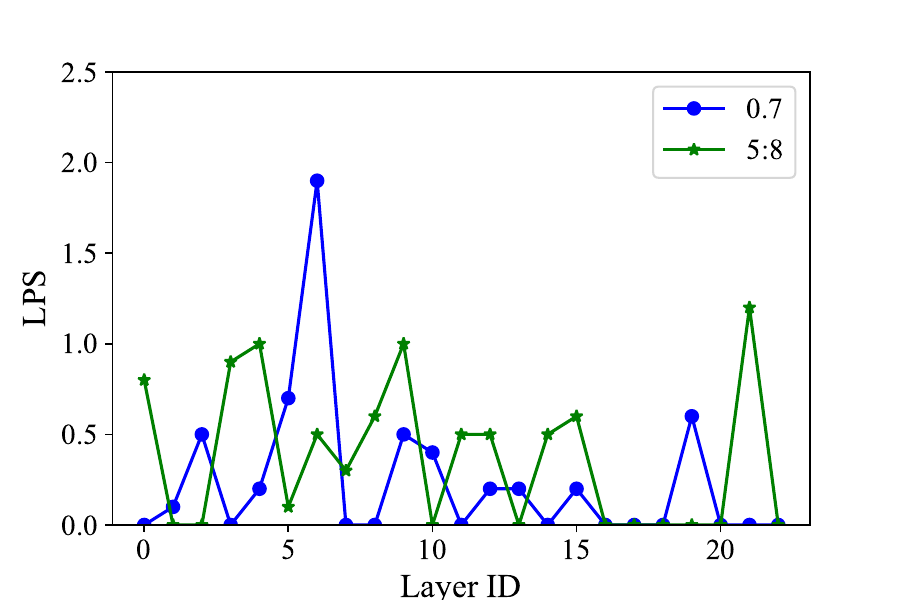}
  \label{OD:2}
  
  \subcaption{LLaVA-7B-Magnitude-Vision} 
  
\end{subfigure}
\caption{The Layerwise Pruning Sensitivity (LPS) of Mixtral-57B and LLaVA-7B at various layerwise sparsity. The results of Mixtral-57B are based on WikiText2 perplexity. The results of LLaVA-7B are based on MM-Vet.}
\vspace{-10pt}
\label{fig3}
\end{figure*}

In addition to the aforementioned standard architectures, we also examine other specialized architectures in Fig. \ref{fig3}, including Mixtral-56B and LLaVA-7B. In both architectures, the layer-wise pruning sensitivity similarly follows a non-uniform distribution. For Mixtral-56B, the sensitivity distribution retains a "U" shape, while for LLaVA-7B, the "U" shape is less pronounced but still exhibits a non-uniform distribution across both the language and vision heads.\par
Additionally, we conducted LPS experiments on multiple zero-shot tasks, including BoolQ \cite{clark2019boolq}, RTE \cite{wang2018glue}, HellaSwag \cite{zimmer2023perp}, WinoGrande \cite{sakaguchi2021winogrande}, ARC Easy and Challenge \cite{clark2018think}, and OpenBookQA \cite{mihaylov2018can}. The results are shown in Table \ref{tab1}. As shown in the table, the sensitivity to pruning varies significantly across different layers on any given dataset.

\subsection{Supplementary Results for Empirical Study 2}

\begin{figure*}
\vskip -0.1in
\centering

\begin{subfigure}[b]{0.45\textwidth}
  \centering
  \includegraphics[width=1\textwidth]{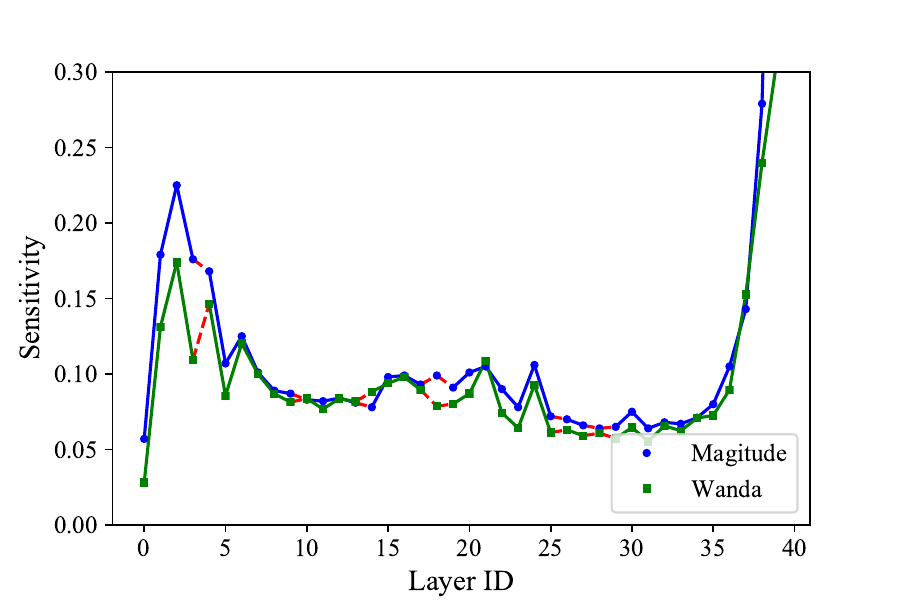}
  \label{OD:1}
  \subcaption{LLaMA2-13B} 
  
\end{subfigure}
\hfill
\begin{subfigure}[b]{0.45\textwidth}
  \centering
  \includegraphics[width=1\textwidth]{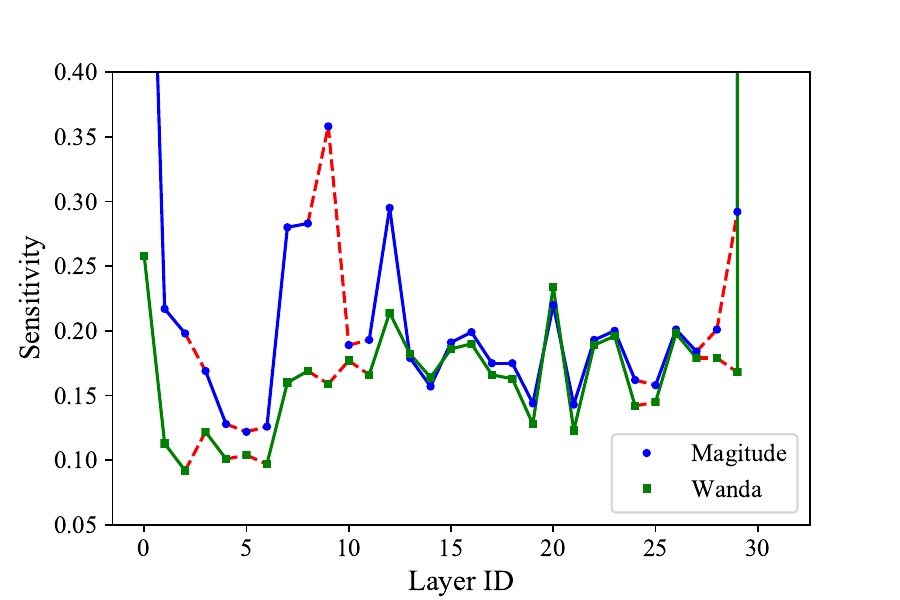}
  \label{OD:2}
  
  \subcaption{Baichuan-7B} 
  
\end{subfigure}
\caption{The LPS (WikiText2 perplexity) of the model under different pruning metrics. The red line highlights layers where sensitivity
reversal occurs, marking only adjacent layers for clarity.}
\vspace{-10pt}
\label{fig4}
\end{figure*}

Fig. \ref{fig4} presents the LPS of more models under different pruning metrics. The phenomenon of sensitivity reversal frequently occurs across these models, supporting the principle that layer-wise sparsity rates should depend on the pruning metrics.\par

\subsection{Supplementary Results for Experiments 5.2}

\begin{figure*}
\vskip -0.1in
\centering

\begin{subfigure}[b]{0.45\textwidth}
  \centering
  \includegraphics[width=1\textwidth]{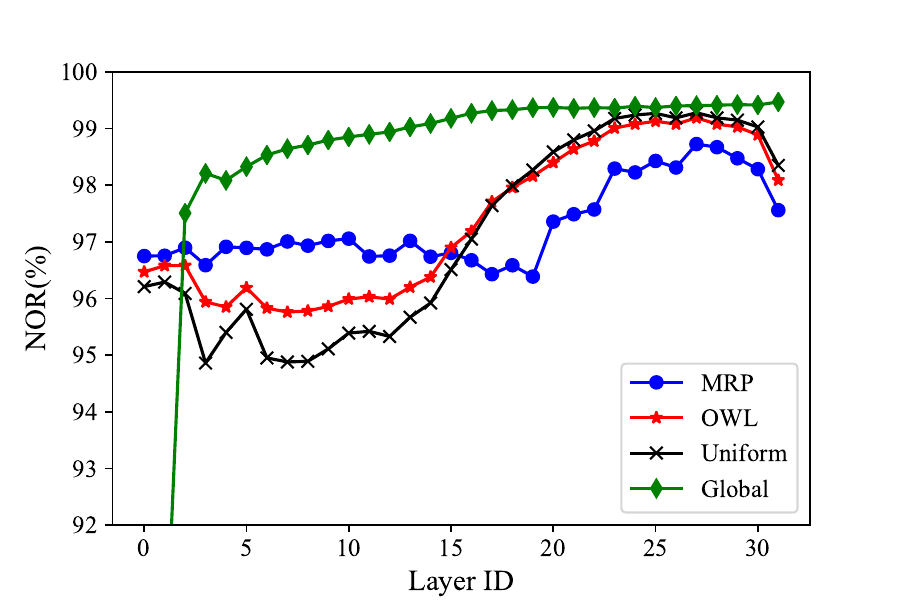}
  \label{OD:1}
  \subcaption{LLaMA2-7B} 
  
\end{subfigure}
\hfill
\begin{subfigure}[b]{0.45\textwidth}
  \centering
  \includegraphics[width=1\textwidth]{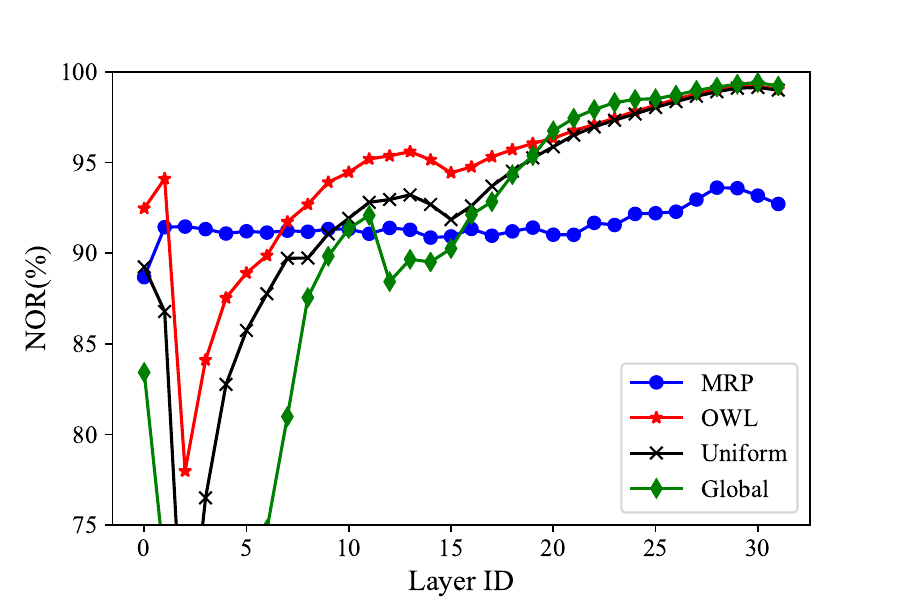}
  \label{OD:2}
  
  \subcaption{OPT-6.7B} 
  
\end{subfigure}
\caption{The non-uniform ratio (NOR) of the model using Wanda pruning.}
\vspace{-10pt}
\label{fig5}
\end{figure*}

We present the layerwise redundancy of sparse
models pruned by different methods, quantified using the Non-Outlier Ratio (NOR) metric, as shown in Fig. \ref{fig5}.  The table shows that on both LLaMA and OPT, the sparse models obtained through MRP exhibit more evenly distributed redundancy across layers.\par
\begin{table*}[htp]

    \centering
    \begin{tabular}{c|c|ccccccccc}
        \hline
        Model  & Metric & Layer ID & BoolQ	&RTE	&HellaSwag	&WinoGrande	&ARC-e	&ARC-c	&OBQA	&Average \\
 
        \hline
        \multirow{8}{*}{LLaMA2-7b} & \multirow{4}{*}{Magnitude} & 0 &1.26	&1.09	&0.29&	0&	0	&0	&0	&0.38 \\
         &  & 2 &3.1	&1.81	&2.63	&0.16	&2.4	&0.94	&0	&1.58\\
         &  & 10 &4.32	&2.53	&1.33	&0.95	&1.64	&0.43	&0	&1.60\\
         &  & 31 &16.76	&9.03	&3.09	&3.16	&6.95	&2.65	&0.6	&6.03\\

        \cline{2-11}
        & \multirow{4}{*}{Wanda} & 0 &0.52	&0	&0.2	&0	&0.34	&0	&0.2	&0.18\\
         &  & 2 &2.17	&2.17	&0.25	&0.31	&1.31	&0.26	&0.2	&0.95\\
         &  & 10 &3.03	&3.61	&1.56	&0.79	&0.72	&0	&0	&1.39\\
         &  & 31 &0.74	&10.11	&1.42	&0	&1.69	&0.51	&0.6	&2.15\\
         \hline
        \multirow{8}{*}{LLaMA2-13b} & \multirow{4}{*}{Magnitude} & 0 &0.67	&3.25	&0.68	&0.24	&0	&0.6	&0.4	&0.83\\
         &  & 2 &0	&3.61	&0.14	&0	&0.04	&1.11	&0.60	&0.79\\
         &  & 10 &0.89	&1.44	&0.4	&0.79	&1.09	&1.70	&1.4	&1.10\\
         &  & 39 &5.9	&0	&1.65	&0.87	&2.61	&1.79	&0.60	&1.92\\

        \cline{2-11}
        & \multirow{4}{*}{Wanda} & 0 &0.18	&0.36	&0.27	&0	&0	&0	&0.40	&0.17\\
         &  & 2 &0	&2.52	&0	&0	&0.29	&0.77	&0	&0.51\\
         &  & 10 &0.43	&1.80	&0.65	&0.55	&0.67	&0.51	&0.20	&0.69\\
         &  & 39 &0	&3.25	&1.13	&0.95	&0	&1.11	&0.20	&0.95\\
         \hline
        \multirow{8}{*}{OPT-6.7b} & \multirow{4}{*}{Magnitude} & 0 &0.59	&0.72	&0	&0	&0.42	&1.2	&0	&0.42\\
         &  & 2 &0.95	&4.69	&4.71	&6.08	&2.53	&1.88	&4.6	&3.63\\
         &  & 10 &0.31	&0	&0.05	&0.39	&0	&0.26	&0.8	&0.26\\
         &  & 31 &12.02	&2.52	&4.08	&0.94	&5.26	&1.79	&1	&3.94\\

        \cline{2-11}
        & \multirow{4}{*}{Wanda} & 0 &0.77	&0.72	&0.19	&0.71	&0	&0.43	&0	&0.40\\
         &  & 2 &4.26	&3.97	&27.47	&15.7	&16.08	&11.26	&10.8	&12.79\\
         &  & 10 &0.25	&0	&0.01	&0.31	&0	&1.03	&0.4	&0.29\\
         &  & 31 &0	&0.72	&1.56	&0.71	&0.13	&1.45	&0.8	&0.77
\\
         \hline
    \end{tabular}
    \caption{Sensitivity of each layer on multiple zero-shot tasks. 0 means no performance degradation.}
    \label{tab1}
\end{table*}
\section{Implementation details}
\subsection{Models and Dataset}
We evaluate MRP's performance across a variety of LLMs, including the LLaMA2 model family (ranging from 7 billion to 70 billion parameters), LLaMA3-8B and OPT-6.7B. Our evaluation follows established LLM pruning methodologies, assessing both language modeling performance and zero-shot capabilities of sparse LLMs. Specifically, we use the Perplexity metric on the WikiText2 validation dataset to measure language modeling performance, and the Accuracy metric for zero-shot evaluations on seven commonsense benchmarks: BoolQ \cite{clark2019boolq}, RTE \cite{wang2018glue}, HellaSwag \cite{zimmer2023perp}, WinoGrande \cite{sakaguchi2021winogrande}, ARC Easy and Challenge \cite{clark2018think}, and OpenBookQA \cite{mihaylov2018can}.
\subsection{Baselines}
Here, we selected three pruning metrics: the traditional Magnitude and the novel Wanda and SparseGPT. We incorporate MRP directly into three metrics. The only distinction between these variants lies in their layerwise sparsity ratios. To evaluate whether MRP can accurately allocate layer-wise sparsity, we compared it with other layer-wise sparsity allocation methods, including Uniform, DSA, and BESA.
\subsection{Hyperparameters}
In this section, we share the hyperparameters used to reproduce the results in our experiments in Table \ref{tab2}.
\begin{table}[h]

    \centering
    \begin{tabular}{ccccc}
        \hline
Model & $r$&$s_0$ & $s_{min}$ & $\alpha$\\
\hline
LLaMA2-7B & 0.50 & 0.20 &0.05& 0.95\\
LLaMA2-13B & 0.55 & 0.20 &0.05& 0.95\\
LLaMA2-70B & 0.65 & 0.06 &0.03& 0.95\\
LLaMA3-8B & 0.60 & 0.15 &0.05& 0.95\\
OPT-6.7B & 0.55 & 0.10 &0.05& 0.95\\
\hline
    \end{tabular}
    \caption{Hyperparameters used to obtain the results in this paper.}
    \label{tab2}
\end{table}
\newpage

\end{document}